\documentclass[10pt,twocolumn,letterpaper]{article}

\usepackage[pagenumbers]{cvpr} 
\usepackage{graphicx}
\usepackage{amsmath}
\usepackage{amssymb}
\usepackage{booktabs}
\newcommand{\ourmodel}{SIA\xspace}

\usepackage{bm}
\usepackage{algpseudocode}
\usepackage{algorithm}
\usepackage{makecell}
\usepackage{amsmath}
\usepackage{multirow}   \usepackage{xspace}     \usepackage{graphicx}
\usepackage{adjustbox}
\usepackage{caption}
\usepackage{subcaption}
\usepackage[dvipsnames]{xcolor}

\definecolor{citecolor}{HTML}{0071bc}
\usepackage[breaklinks,colorlinks,citecolor=citecolor,bookmarks=false]{hyperref}

\usepackage[capitalize]{cleveref}
\crefname{section}{Sec.}{Secs.}
\Crefname{section}{Section}{Sections}
\Crefname{table}{Table}{Tables}
\crefname{table}{Tab.}{Tabs.}

\begin{document}

\title{Semantic Image Attack for Visual Model Diagnosis}
\author{Jinqi Luo \quad Zhaoning Wang \quad Chen Henry Wu \quad Dong Huang \quad Fernando De la Torre \\
Robotics Institute, Carnegie Mellon University, Pittsburgh, PA, USA\\
{\tt\small \{jinqil, zhaoning, chenwu2, dghuang, ftorre\}@cs.cmu.edu}
}

\maketitle
\vspace{-3mm}
\begin{abstract}
In practice, metric analysis on a specific train and test dataset does not guarantee reliable or fair ML models. This is partially due to the fact that obtaining a balanced, diverse, and perfectly labeled dataset is typically expensive, time-consuming, and error-prone. Rather than relying on a carefully designed test set to assess ML models' failures, fairness, or robustness, this paper proposes Semantic Image Attack (\ourmodel), a method based on the adversarial attack that provides semantic adversarial images to allow model diagnosis, interpretability, and robustness. Traditional adversarial training is a popular methodology for robustifying ML models against attacks. However, existing adversarial methods do not combine the two aspects that enable the interpretation and analysis of the model's flaws: semantic traceability and perceptual quality. \ourmodel combines the two features via iterative gradient ascent on a predefined semantic attribute space and the image space. We illustrate the validity of our approach in three scenarios for keypoint detection and classification.  (1) Model diagnosis: \ourmodel generates a histogram of attributes that highlights the semantic vulnerability of the ML model (i.e., attributes that make the model fail). (2) Stronger attacks: \ourmodel generates adversarial examples with visually interpretable attributes that lead to higher attack success rates than baseline methods. The adversarial training on \ourmodel improves the transferable robustness across different gradient-based attacks. (3) Robustness to imbalanced datasets: we use \ourmodel to augment the underrepresented classes, which outperforms strong augmentation and re-balancing baselines. \footnote{This paper was first submitted to NeurIPS on May 9, 2022.}
\end{abstract}

\begin{figure}
    \centering
    \includegraphics[width=\linewidth]{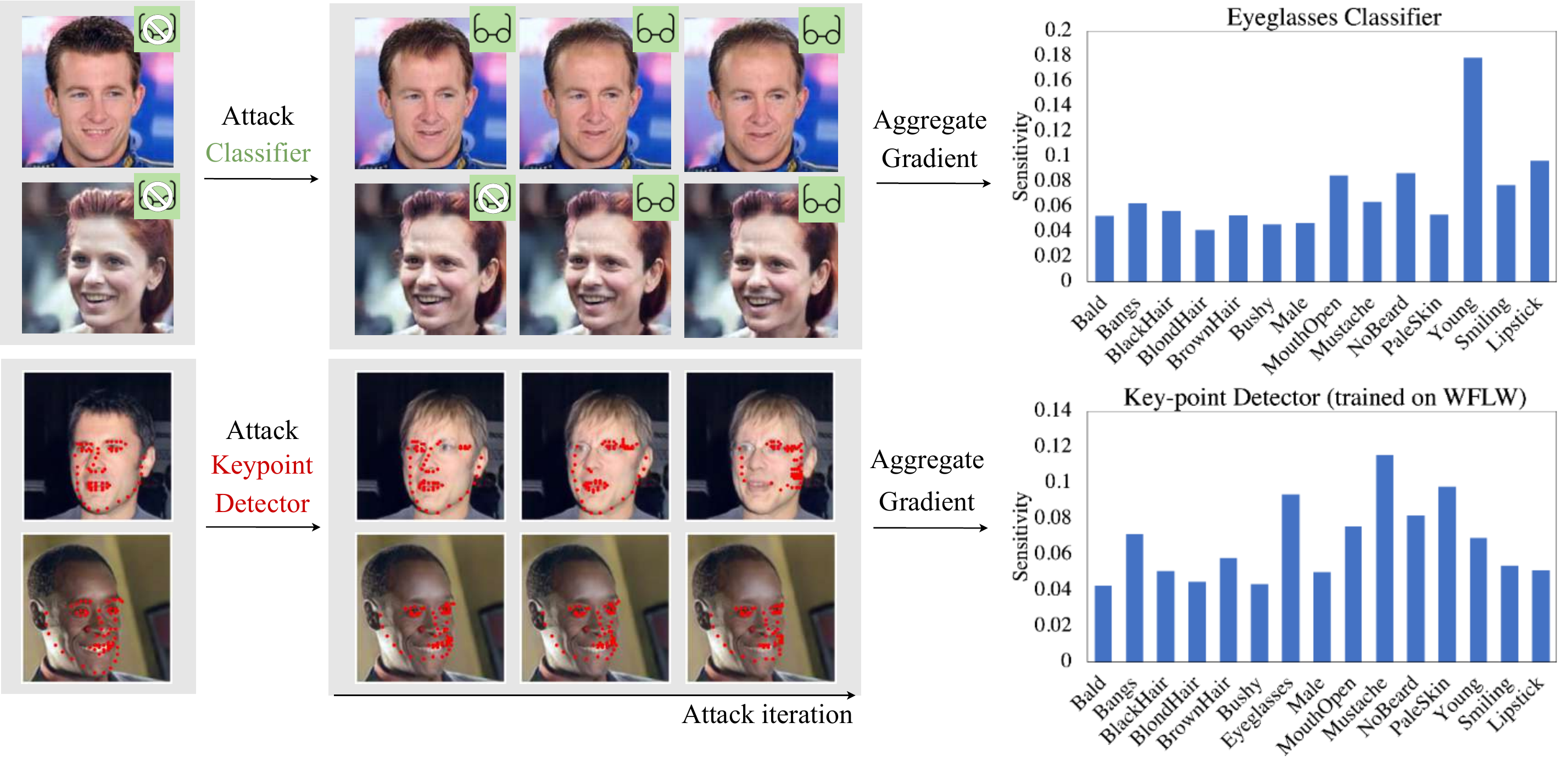}
    \caption{Model diagnosis by \ourmodel. 
    The models to be diagnosed are an eyeglasses classifier (top two rows) and a keypoint detector (bottom two rows). \ourmodel reveals that the eyeglasses classifier is more sensitive to lipstick and age, whereas the keypoint detection tends to fail on people with moustache and pale skin. See text for an explanation of the figure.    }
    \label{fig:demo_iteration}
    \vspace{-3mm}
\end{figure}

\section{Introduction}

In Machine Learning (ML), error analysis of train and test data is a critical stage in model assessment and debugging. However, the conclusions extracted from the metric analysis on the train or test data do not guarantee reliability nor fairness, partially due to the fact that datasets are imperfect \cite{Sattigeri2018FairnessG,Ramaswamy2021FairAC}.  Even with careful collection and filtering, data naturally contain biases. Furthermore, in the case of computer vision learning systems, having a uniform distribution over all conceivable variability of an object in an image (e.g., position, lighting, background) is typically impractical (i.e., exponential) and labels are prone to errors. The issue only grows more severe with large-scale datasets. ML models trained on these datasets inevitably inherit these imbalances and biases. These limitations also apply to test sets that are typically used for model evaluation. 
Such a vulnerability is a landmine that must be recognized and processed in order for ML applications to succeed.  The question we strive to address in this study is whether there are alternative/better methods for discovering biases and performing model diagnostics in computer vision models instead of only relying on a test set.

Fig.~\ref{fig:demo_iteration} illustrates the problems that this paper tries to address. Given an eyeglasses classifier (top two rows) or a keypoint detector (bottom two rows), which kind of face images will lead to misclassification or misdetection? How can we automatically discover these failure cases and robustify the model? How can we perform visual model diagnosis in a semantic attribute space?  To accomplish these, we propose Semantic Image Attack (\ourmodel), a new adversarial attack in a generative model of faces parameterized by attributes. In top left in Fig.~\ref{fig:demo_iteration}, we see two images of faces without eyeglasses, and the model classifies them correctly. After several iterations of \ourmodel (right column), our model is able to modify facial attributes (e.g., smile, eye color, facial hair) to mislead the eyeglass classifier. Also, our model builds a histogram of the sensitivity across attributes (i.e., visual model diagnosis). While evaluating the model resilience on a single attribute can be relatively straightforward, evaluating the model robustness for combinations of attributes can be quite challenging (due to the exponential nature of attribute combinations). \ourmodel is able to {\em jointly} search over the space of attributes, and hence performs a multi-attribute attack for model diagnosis.  Similarly, in Fig.~\ref{fig:demo_iteration}, the bottom two rows illustrate the model diagnosis results for keypoint detection.

\begin{figure*}
\vspace{-2mm}
    \centering
    \includegraphics[width=0.88\textwidth]{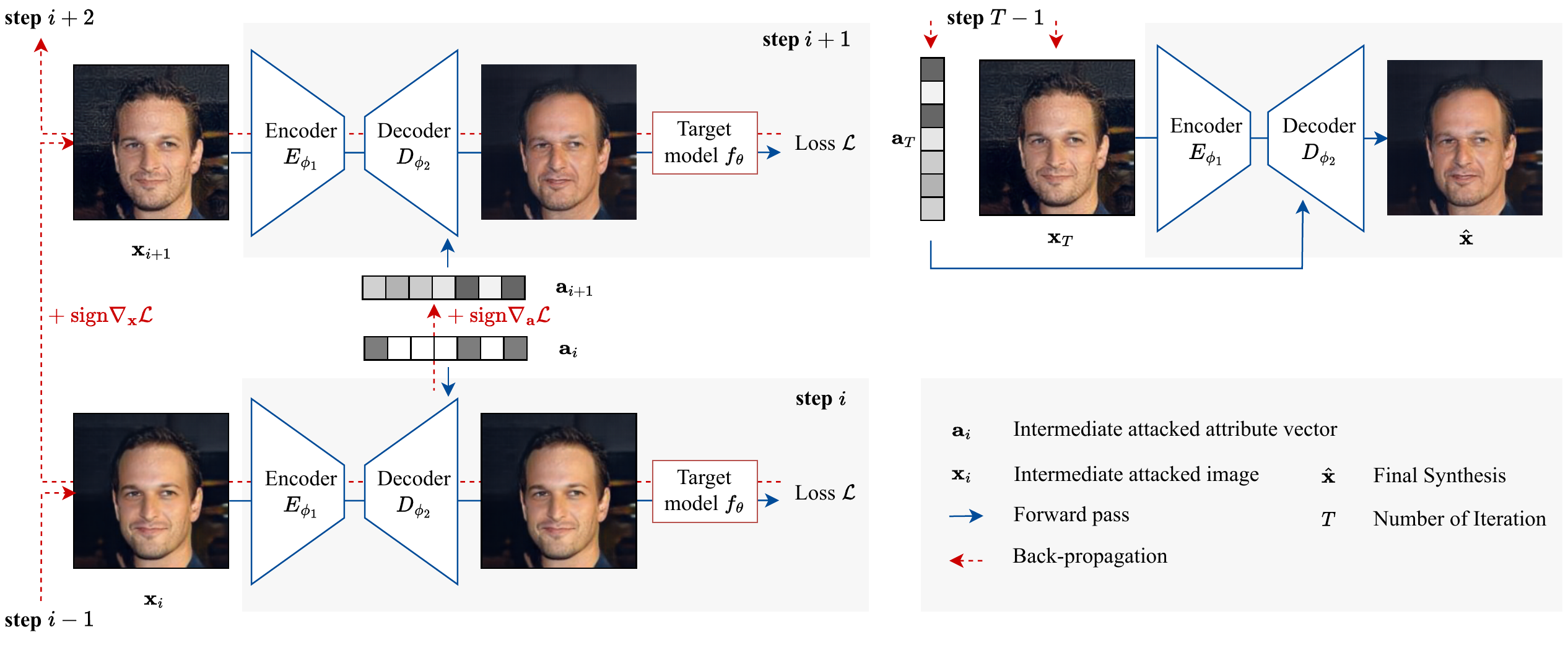}
    \caption{The \ourmodel framework uses an encoder-decoder GAN $\mathcal{G}_\phi = \{E_{\phi_{1}}$, $D_{\phi_{2}}\}$ to attack a target model $f_\theta$.
In each iteration $i$, we update the image $\mathbf{x}_i$ and the attribute vector $\mathbf{a}_i$ using the gradients from the loss $\mathcal{L}$ (see Eq.~(\ref{eq:gradient-ascent-attribute})). Finally, the encoder-decoder GAN projects the attacked image $\mathbf{x}_T$ and attributes $\mathbf{a}_T$ in the last iteration back to the image manifold to produce adversary $\mathbf{\hat{x}} = \mathcal{G}_\phi(\mathbf{x}_T, \mathbf{a}_T)$. Solid lines stand for forward passes, and dashed lines stand for backpropagation. }
    \label{fig:model}
    \vspace{-3mm}
\end{figure*}

In addition to model diagnosis, \ourmodel is able to
robustify the target model by re-training the model on adversarial examples (see Fig.~\ref{fig:demo_iteration} middle columns). In our experiments, we also show the robustness from \ourmodel is more transferable to other types of attacks than other competing attack methods.
Finally, we show that \ourmodel outperforms popular image augmentation techniques \cite{Cubuk_2019_CVPR_autoaugment,2019cutmix} and re-balancing baselines when learning from imbalanced datasets.

\section{Related Work}

\subsection{Adversarial Attacks} Gradient-guided image space perturbation attacks have been popular in robustifying ML models~\cite{GoodfellowSS14,madry2018towards}. The image perturbations generated by such attacks are small image changes typically imperceptible to humans. \cite{Zhu_2021_ICCV, Miss_the_Point} adopted such attacks on keypoint detectors to robustify detectors against adversarial perturbations. \cite{ijcai2018-543} was pioneering in using Generative Adversarial Networks (GANs) \cite{NIPS2014_5ca3e9b1} to generate adversarial attacks. However, \cite{ijcai2018-543} only allowed a limited perturbation bound and required individually trained GANs for every target ML model. A major issue of previous methods is the lack of interpretability of the attack. To address this issue, \cite{2020semanticadv} used the interpolation of semantic feature maps to generate attacks, and showed the effectiveness in terms of the attack's success rate in classification and detection problems.  \cite{Gokhale2021AttributeGuidedAT} also modeled the perturbations in the attribute space, and showed that the attribute space can improve model robustness. However, this work aims to find perturbations in samples that do not change labels, and their model is not robust to small perturbation attacks in the image space.  Moreover,  \cite{Gokhale2021AttributeGuidedAT} did not provide interpretability into the failures of the computer vision model. Similarly, \cite{Li_2021_CVPR} sampled images in the latent space of a GAN to generate strong attacks, but their attacks are not interpretable in the attribute space. \cite{2019semanticadversarial} conducted model attacks only in the attribute space using the attribute-assisted GAN (AttGAN) \cite{2019attgan}. This approach does not attack the image space and does not constrain the scale of parametric gradients, which leads to generating unrealistic images.

Unlike previous work in the adversarial attack literature, \ourmodel performs gradient-guided attack simultaneously in the image and a pre-defined attribute space. As we will show in the experimental section, performing gradient ascent only in the attribute space leads to unstable results. In addition, our approach only uses \textit{one} GAN backbone \cite{2019attgan} to attack all target models (i.e., AttGAN can be used to evaluate any computer vision model). Finally, our method provides a histogram of the sensitivity of the target models across attributes of interest. This information can be critical to gather insights into the fairness and robustness of the model. 

\subsection{Bias and Fairness Analysis}  
\cite{Balakrishnan2020TowardsCB, Denton2019ImageCS} showed that by traversing images in the GAN latent space, one can visualize the attribute-wise sensitivity of a target classifier. But such a process requires manual annotation of the generated images, which is expensive and infeasible for large attribute spaces. Recently, \cite{Lang_2021_ICCV} used StyleGAN~\cite{Karras2019ASG} to learn a target-model-dependent style/attribute space, which allows a human to interpret the target models' behavior in terms of attributes. Furthermore, several previous works proposed fairness metrics to evaluate a model without a fair test set \cite{Hardt2016EqualityOO,Ramaswamy2021FairAC,2018Mitigating}.  While previous fairness metrics focus on a model's statistical behavior across attributes,
\ourmodel focuses on the model's decision for each instance (though individual sensitivities can be further aggregated to get sub-population sensitivity, see Fig.~\ref{fig:demo_iteration}). Moreover, \ourmodel is able to search over attribute combinations. 

\section{Semantic Image Attack (\ourmodel)}

This section describes our \ourmodel algorithm starting with the notation.  

{\bf Target model ($f_\theta$):} Let $f_\theta$, parameterized by $\theta$, be the target model that we want to improve or perform model diagnosis on. In this paper, we cover two types of neural network models $f_\theta$: an attribute classifier and a keypoint detector. 

An attribute classifier takes an image $\mathbf{x}$ as input and outputs $f_\theta(\mathbf{a}|\mathbf{x})$, the conditional probability of attribute $\mathbf{a} \in \mathcal{A}$ given $\mathbf{x}$, where $\mathcal{A}$ is the attribute space. Without loss of generality, we consider binary classifiers. Given the ground truth class label $c$ of the image $\mathbf{x}$, the classification loss is defined as the binary cross-entropy $\mathcal{L_{\theta}} =  -(c\log{f_\theta(c|\mathbf{x})} + (1-c)(\log{(1-f_\theta(c|\mathbf{x}))}))$. 

The keypoint detector takes an image $\mathbf{x}$ as input and outputs $f_\theta(\mathbf{p}|\mathbf{x})$, the
probability heatmap of the keypoints $\mathbf{p} \in \mathcal{P}$, where $\mathcal{P}$ is the 2D pixel coordinate space. Given a training image $\mathbf{x}$ with ground truth facial keypoints $\mathbf{c}$, the loss $\mathcal{L_{\theta}}$ is defined as the mean squared error between the predicted heatmap and the ground-truth heatmap corresponding to $\mathbf{c}$, see \cite{WangSCJDZLMTWLX19} for details.

{\bf Adversary ($\mathbf{\hat{x}}$):} For each input image $\mathbf{x}$, an adversarial example $\mathbf{\hat{x}}$ is a synthesized image that misleads the target model $f_\theta$ to produce outputs that are far away from the ground truth $\mathbf{c}$ or changes the label of the classifier. Different from traditional adversarial attack methods, SIA generates adversarial examples under a combination of perturbations in the attribute and image spaces. 

\ourmodel consists of two main components: (1) an AttGAN $\mathcal{G}_\phi = \{E_{\phi_{1}}$, $D_{\phi_{2}}\}$,
$    \mathcal{G}_\phi(\mathbf{x}, \mathbf{a}) = D_{\phi_{2}}([E_{\phi_{1}}(\mathbf{x}); \mathbf{a}])$,
where the encoder $E_{\phi_{1}}$ maps an input image $\mathbf{x}$ to a latent vector, the decoder $D_{\phi_{2}}$  takes as an input the concatenation of $E_{\phi_{1}}(\mathbf{x})$ and the attribute vector $\mathbf{a}$ to generate an image; (2) a pretrained target model $f_\theta$ to be diagnosed.

\subsection{Generating Iterative Adversaries}
Our framework uses both the attribute space and the image space to iteratively generate adversarial images $\mathbf{\hat{x}}$.  
We iteratively compute gradient ascent in the attribute space and the image space.  An advantage of optimizing over the attribute and image space is an improved adversarial space, that leads to a better generation of adversarial examples (see experiment section). 

The procedure to {\em jointly} update the attribute vectors and images is as follows:  \begin{equation}
\label{eq:gradient-ascent-attribute}
\footnotesize
\begin{aligned}
    \mathbf{a}_{i} &= \Pi_{\mathcal{B}(\epsilon_{\mathbf{a}})}(\mathbf{a}_{i-1} + \eta \operatorname{sign} [\nabla_{\mathbf{a}} (\mathcal{L_{\theta}} (f_\theta(\mathcal{G}_\phi(\mathbf{x}_{i-1}, \mathbf{a}_{i-1}))))]), \\
    \mathbf{x}_{i} &= \Pi_{\mathcal{B}(\epsilon_{\mathbf{x}})}(\mathbf{x}_{i-1} + \eta \operatorname{sign} [\nabla_{\mathbf{x}} (\mathcal{L_{\theta}}(f_\theta(\mathcal{G}_\phi(\mathbf{x}_{i-1}, \mathbf{a}_{i-1}))))]).
\end{aligned}
\end{equation}
The adversarial example $\mathbf{\hat{x}}$ is an image space projection of a fine-grained perturbation of the original input image $\mathbf{x}$ at both  pixel  and attribute levels. During the process, our \ourmodel framework manipulates the attribute vector in a predefined attribute space such that the target model is compromised. Note that each iteration of the updates will be clipped with a radius $\epsilon$ to make sure that the perturbation is bounded and valid. The pixel-level perturbed image is fed into $\mathcal{G}_\phi$ to encode the adversarial information into the latent vector, which is concatenated with the perturbed attribute vector. Specifically, instead of directly perturbing the output image, which may significantly harm the perceptual quality, we perturb the input attribute and the image and let $\mathcal{G}_\phi$ project the perturbed image and attribute back to the image manifold. To prevent synthesis collapse, we adopt the projection $\Pi$ onto the $\ell_\infty$ ball $\mathcal{B}$ of radius $\epsilon$ to constrain the optimization. The projection to generate the final adversarial example is formulated as  $ \mathbf{\hat{x}} = \mathcal{G}_\phi(\mathbf{x}_T, \mathbf{a}_T)$. An overview of our \ourmodel framework is shown in Figure~\ref{fig:model}.

At this point, it is important to notice that perturbing in both the image space and attribute space produces higher attack success rate and finer visual adversarial images. Also, we do it for a fair comparison with traditional methods. Recall that directly perturbing the semantic space limits the attacking capability. Our hybrid attack gives us the flexibility to analyze both the semantic and pixel-level robustness of the model. In fact, SIA’s pixel-level perturbation helps to avoid exaggerated semantic variation that makes the image generation collapse.  An ablation study that illustrates the advantages of perturbing in both the image and attribute space is included in the experimental section. 

\subsection{Interpreting and Improving the Target Model}
Given a set of image-attribute pairs $(\mathbf{x}^{(p)}, \mathbf{a}^{(p)})$  ($p=1,\dots, N$), we run $T$ iterations of Eq. \ref{eq:gradient-ascent-attribute} and store all the generated adversaries. By calculating the absolute variation of attributes during the generation of adversaries $\hat{\mathbf{x}}^{(p)}$, we can discover the most sensitive attribute(s) to the target model $f_\theta(\cdot)$ in the $\mathcal{G}_\phi$'s attribute space. 
We define the sensitivity vector containing sensitivities (in the range of $[0,1]$) of the target model on each attribute as follows: 
\begin{align}
\label{eq:diagnose}
        \mathbf{s} = \frac{1}{N}\sum_{p=1}^N (|\mathbf{a}_{T}^{(p)} - \mathbf{a}_{1}^{(p)}|),
\end{align}
Each value in $\mathbf{s}$ will represent the average perturbation of the corresponding attribute across all sampled images. 
Note that this method can be extended to select top-k attributes that have a greater influence on the prediction of the target model. 
The generated adversaries $\hat{\mathbf{x}}^{(p)}$ are associated with more diverse attribute vectors $\hat{\mathbf{a}}$, which can be considered as an augmented dataset for adversarial training.  See Algorithm~\ref{alg:framework} for more details on how to generate adversaries and sensitivity analysis. 
\begin{algorithm}[!h]
\small
\caption{SIA to generate adversarial examples and sensitivity analysis. 
}\label{alg:framework}
\begin{algorithmic}
\State \textbf{Input: } A set of image-attribute pairs $\{(\mathbf{x}_{0}^{(p)}, \mathbf{a}_{0}^{(p)})\}_{p=1}^{N}$; target model $f_\theta(\cdot)$ 
\State \textbf{Output: } Model sensitivity $\mathbf{s}$; a set of adversaries $\{\hat{\mathbf{x}}^{(p)}\}_{p=1}^{N}$ \For{$p \in \{1, \ldots, N\}$}
    \For{$i \in \{1, \ldots, T\}$} 
        \State $\mathbf{a}_{i}^{(p)} \gets \mathbf{a}_{i-1}^{(p)} + \eta \operatorname{sign} [\nabla_{\mathbf{a}} (\mathcal{L_{\theta}} (f_\theta(\mathcal{G}_\phi(\mathbf{x}_{i-1}^{(p)}, \mathbf{a}_{i-1}^{(p)}))))]$
        \State $\mathbf{a}_{i}^{(p)} \gets \Pi_{\mathcal{B}(\epsilon_{\mathbf{a}})}(\mathbf{a}_{i}^{(p)})$
        \State $\mathbf{x}_{i}^{(p)} \gets \mathbf{x}_{i-1}^{(p)} + \eta \operatorname{sign} [\nabla_{\mathbf{x}} (\mathcal{L_{\theta}}(f_\theta(\mathcal{G}_\phi(\mathbf{x}_{i-1}^{(p)}, \mathbf{a}_{i-1}^{(p)}))))]$
        \State $\mathbf{x}_{i}^{(p)} \gets \Pi_{\mathcal{B}(\epsilon_{\mathbf{x}})}(\mathbf{x}_{i}^{(p)})$
                                        \EndFor
    \State $\mathbf{\hat{x}}^{(p)} \gets \mathcal{G}_\phi(\mathbf{x}_T^{(p)}, \mathbf{a}_T^{(p)}) $
\EndFor
    \State $\mathbf{s} = \frac{1}{N}\sum_{p=1}^N (|\mathbf{a}_{T}^{(p)} - \mathbf{a}_{1}^{(p)}|)$ 

\end{algorithmic}
\label{pseudo_algorithm}
\end{algorithm}
\vspace{-3mm}

\section{Experiments}
This section explains the experimental validation to demonstrate the benefits of SIA for visual model diagnostics, improved robustness against visual attacks, and imbalanced robustness.  

\subsection{Experimental Setups} \label{experiment-setup}

{\bf Attribute-assisted GAN:} Our backbone of AttGAN $\mathcal{G}_\phi$ is trained on the whole CelebA dataset~\cite{liu2015faceattributes}, using $15$ attributes\footnote{we used Bald, Bangs, Black\_Hair, Blond\_Hair, Brown\_Hair, Bushy\_Eyebrows, Eyeglasses, Male, Mouth\_Slightly\_Open, Mustache, No\_Beard, Pale\_Skin, Young, Smiling, Wearing\_Lipstick}. Images are center cropped, resized to $(224, 224)$, and normalized using the ImageNet normalization. $\mathcal{G}_\phi$'s encoding and decoding dimensions are both 64. Shortcuts and inject layers are activated, and the Wasserstein loss \cite{Arjovsky2017WassersteinG} is used. 
We used the codes provided by~\cite{2019attgan}\footnote{https://github.com/elvisyjlin/AttGAN-PyTorch}. 

{\bf Attribute Classifier:} Our classifiers are fine-tuned from TorchVision's pre-trained ResNet50. Unless otherwise stated, we trained binary classifiers on the CelebA training set~\cite{liu2015faceattributes}. For training, we used the Adam optimizer with a learning rate of 0.001 and batch size of 128. The seed for random number generation is 42 for Numpy and PyTorch.

{\bf Keypoint Detector:} We used 
the HR-Net architecture~\cite{WangSCJDZLMTWLX19}. We trained two models, one trained on the Wilder Facial Landmark in the Wild (WFLW) dataset~\cite{wayne2018lab} and the other on the Microsoft (Fake-it) synthetic dataset~\cite{wood2021fake}. To train the two keypoint detectors, we used all images ($10,000$) from the WFLW dataset and the first $10,000$ images from the Fake-it dataset, respectively. We trained with $98$ keypoints on the WFLW dataset and $68$ keypoints on the Fake-it dataset.

\begin{figure}[t]
    \centering
    \includegraphics[width=\linewidth]{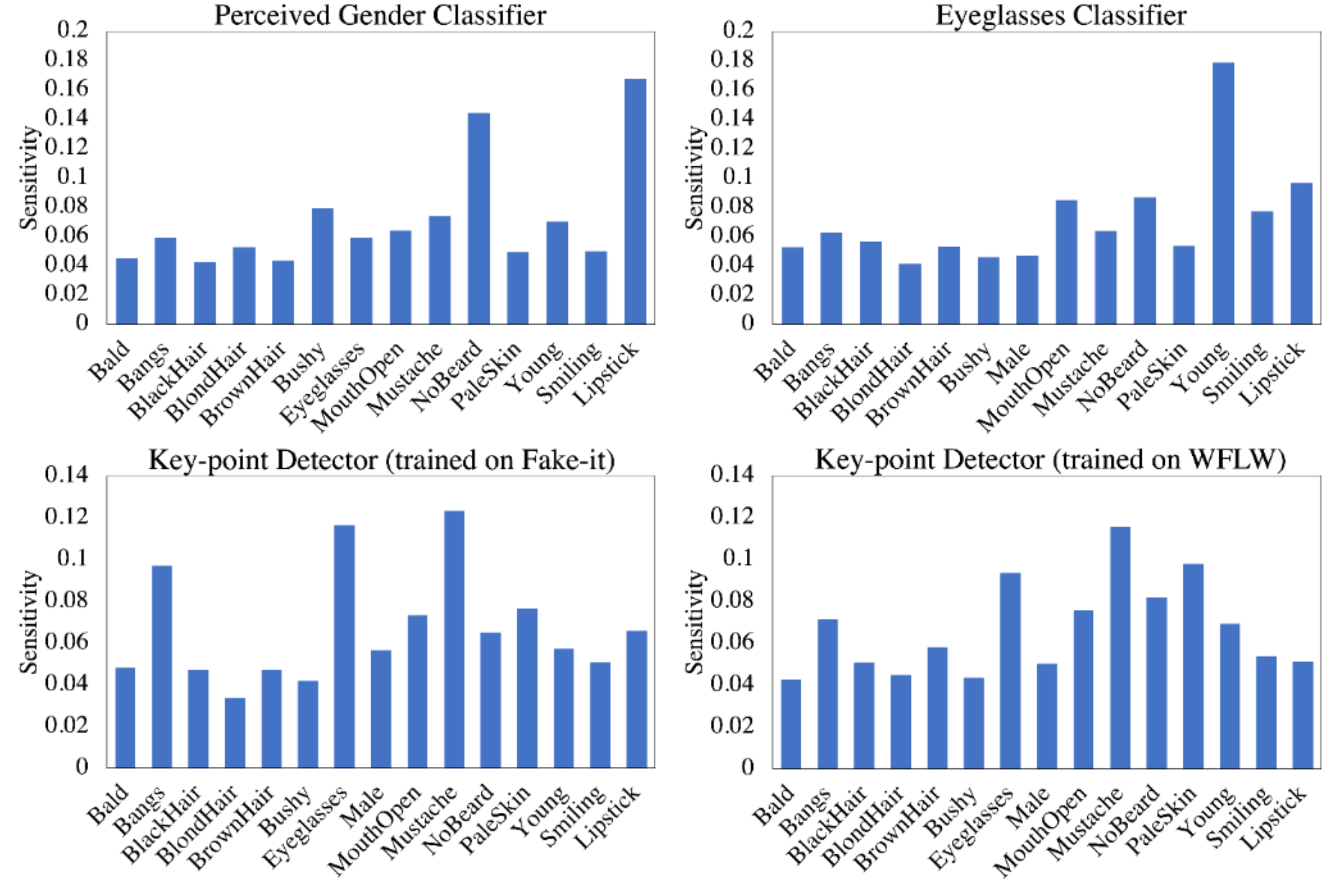}
    \caption{Attribute sensitivity analysis generated by \ourmodel for different classifiers (top) and keypoint detectors (bottom). Perceived gender and eyeglasses classifiers are sensitive to different attributes. However, the keypoint detectors trained on synthetic (left) and real (right) data are sensitive to similar attributes, but the one trained on synthetic data is slightly more sensitive than the one trained on real data.     }
    \label{fig:demo_histogram}
\end{figure}

\begin{figure}[t]
    \begin{subfigure}[b]{\linewidth}
    \centering
    \includegraphics[width=\textwidth]{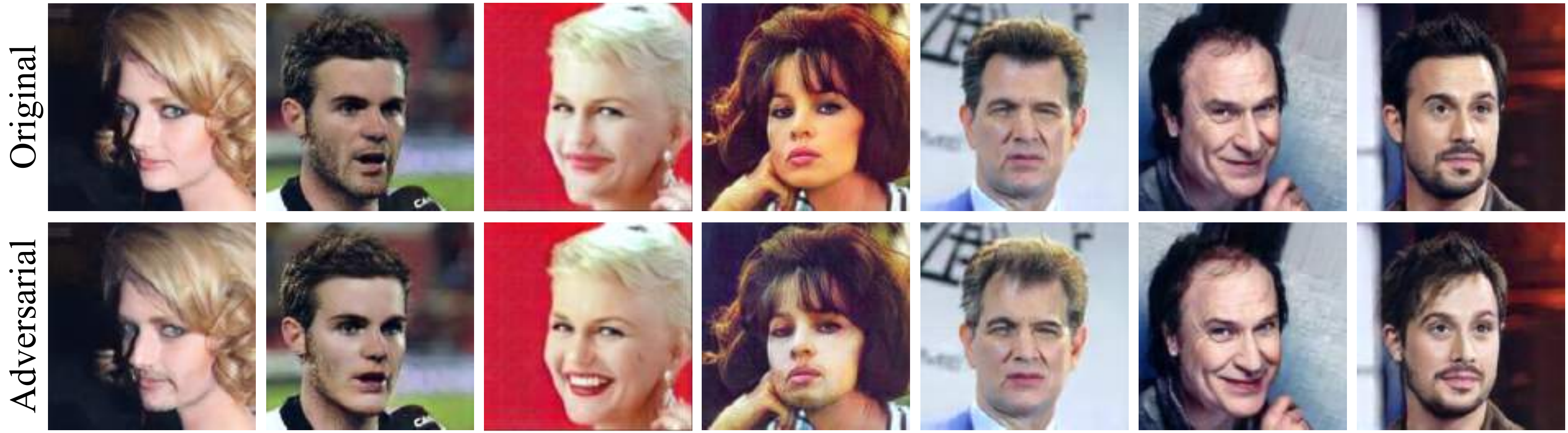}
    \caption{Examples of \ourmodel on perceived-gender classifier}
    \end{subfigure}
    \begin{subfigure}[b]{\linewidth}
    \centering
    \includegraphics[width=\textwidth]{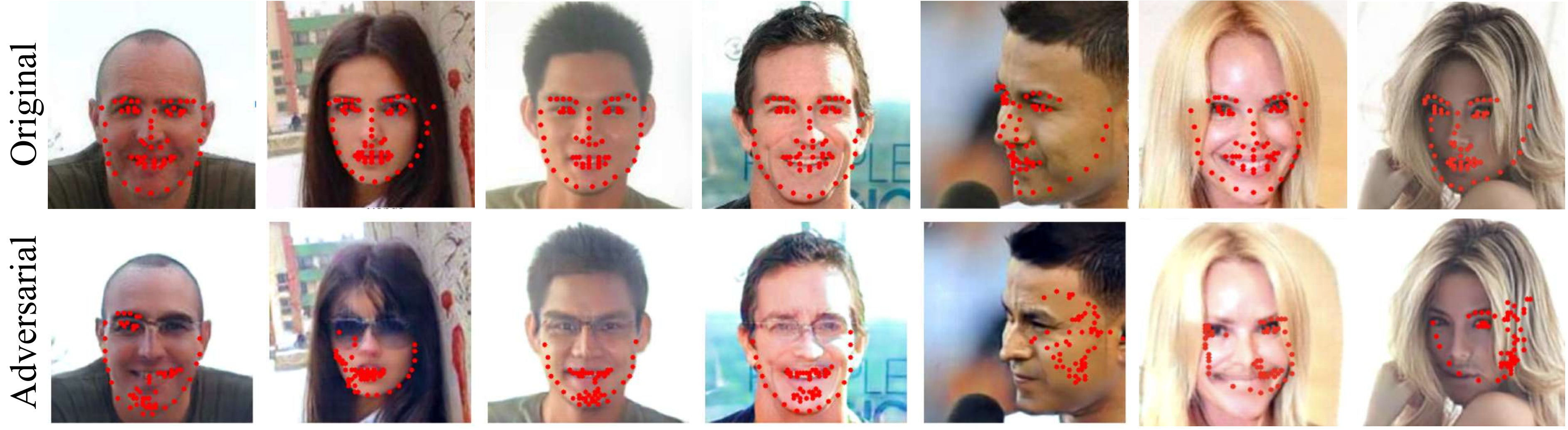}
    \caption{Examples of \ourmodel on Fake-it \cite{wood2021fake} keypoint detector}
    \end{subfigure}
    \vspace{-3mm}
    \caption{ \ourmodel adversarial examples on different target models. 
    }
    \label{fig:demo_images}
    \vspace{-3mm}
\end{figure}

\begin{figure*}[h]
    \vspace{-3mm}
    \begin{subfigure}[b]{0.49\linewidth}
    \centering
    \includegraphics[width=\textwidth]{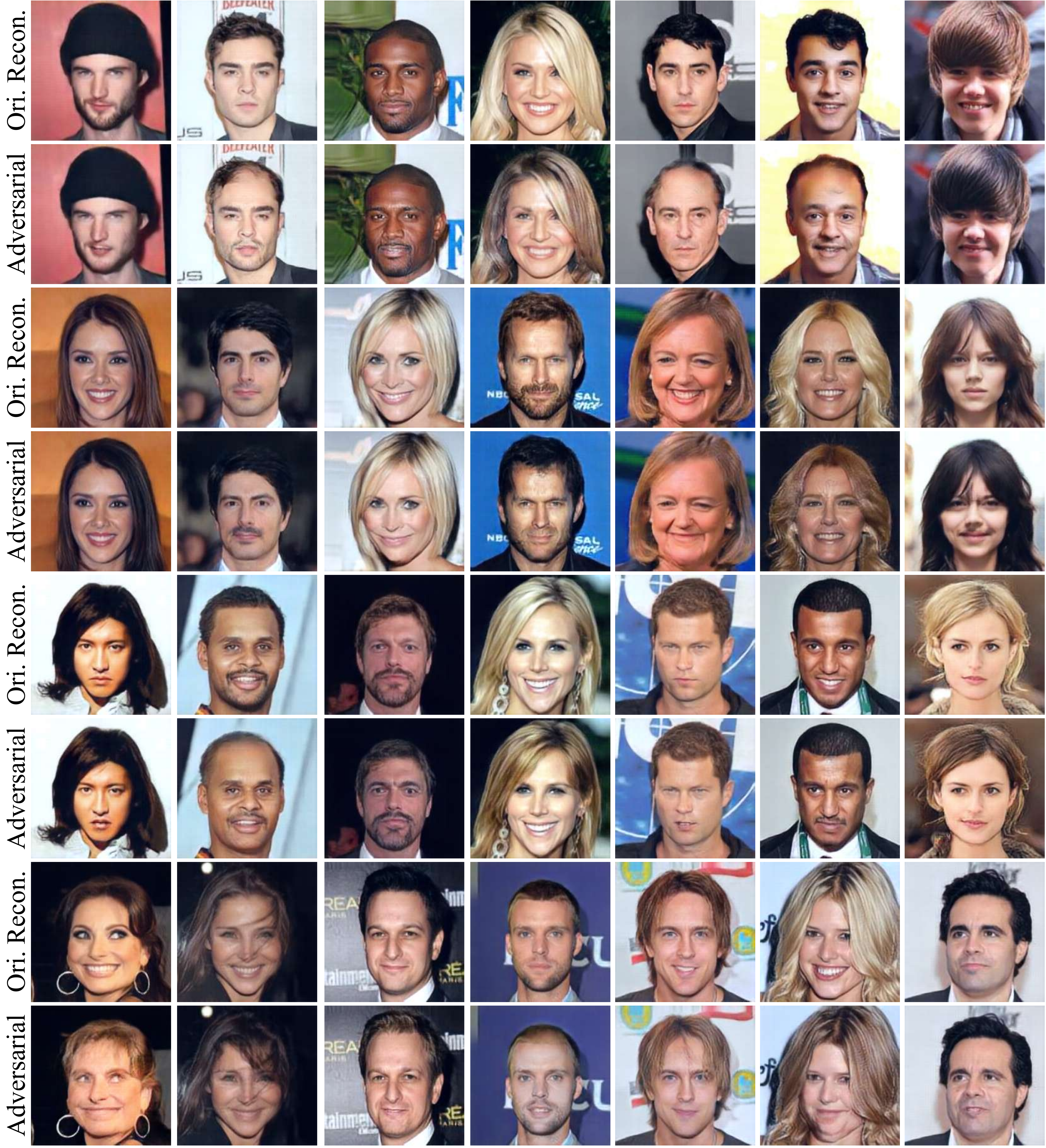}
    \caption{Examples of \ourmodel on eyeglass classifier}
    \end{subfigure}
    \begin{subfigure}[b]{0.49\linewidth}
    \centering
    \includegraphics[width=\textwidth]{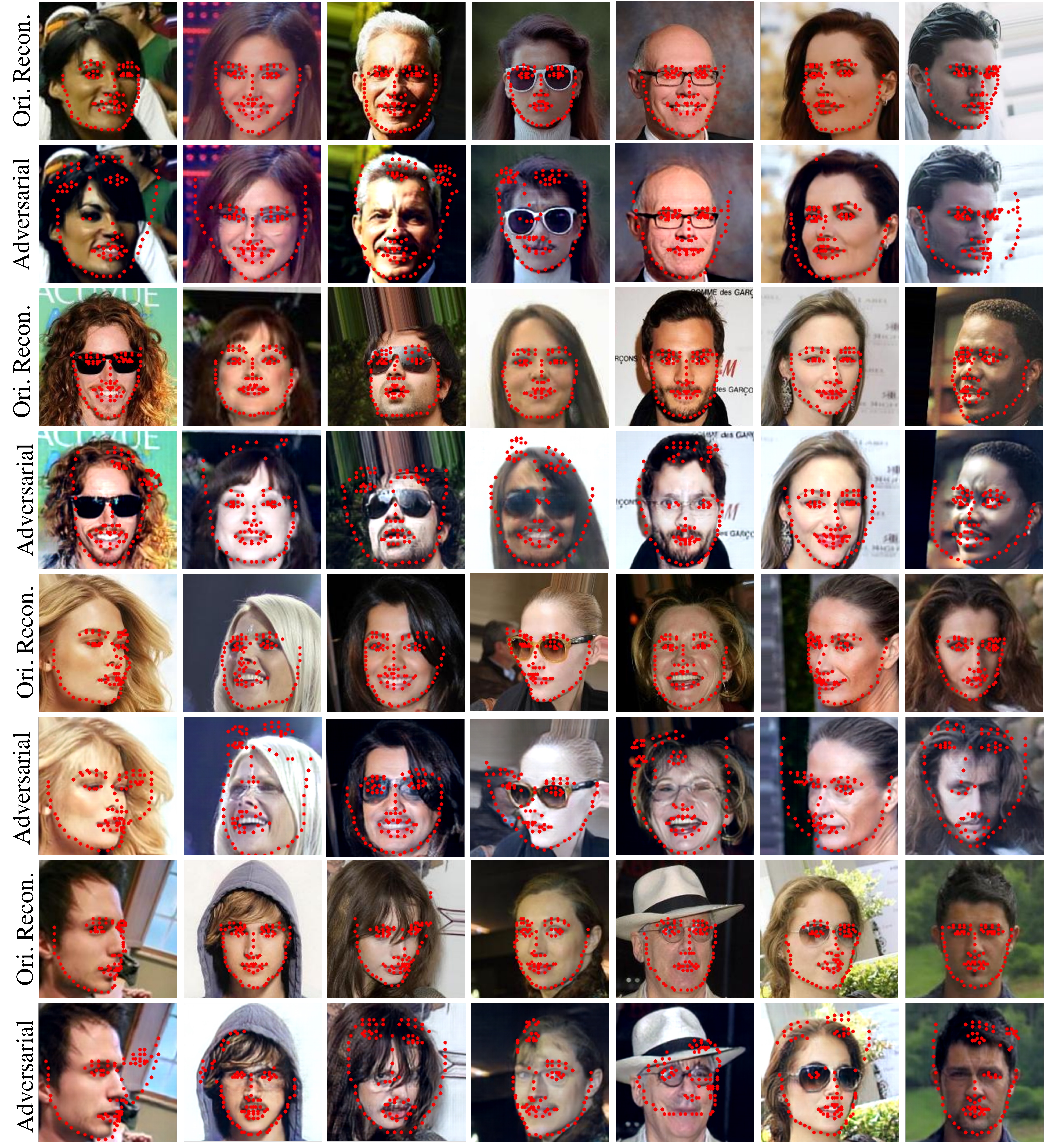}
    \caption{Examples of \ourmodel on WFLW \cite{wayne2018lab} keypoint detector}
    \end{subfigure}
    \vspace{-2mm}
    \caption{(Cont.) \ourmodel adversarial examples on different target models. 
    }
    \label{fig:demo_images_cont}
    \vspace{-4mm}
\end{figure*}

\subsection{Visual Model Diagnosis} \label{visual_model_dianosis}

After training a deep learning model and tuning hyper-parameters of the model on a validation set, an important step is error analysis. The error analysis includes analyzing where the model fails on test data and making systematic changes based on the insights. However, in some scenarios, it is difficult to collect test data across all possible attributes of interest in a uniform manner. Instead of collecting test data, this section describes how SIA can be used for model diagnosis and provides insights into the image attributes that make the model fail.

\subsubsection{Diagnosis visualization}
We trained $8$ binary classifiers on the following attributes from CelebA: \textit{Attractive, Arched\_Eyebrows, Blurry, Chubby, Eyeglasses, Male, No\_Beard, Sideburns} with the setup mentioned in Section 4.1. In addition, we trained two keypoint detection algorithms, one on real images and another one on synthetic images, using the same architecture HR-Net \cite{WangSCJDZLMTWLX19}. \ourmodel reports the sensitivity of the target model w.r.t. different attributes, which is formalized in  Eq.~\ref{eq:diagnose}.
We selected the first $10,000$ images in CelebA to evaluate the sensitivities. Fig.~\ref{fig:demo_histogram} illustrates the histogram for the classifier (first row) and keypoint detector (second row) towards different attributes, according to Eq.~\ref{eq:diagnose}.  For clearer visualization, we have normalized the sensitivity for each attribute by the sum of sensitivities. We can see that for the perceived-gender classifier, lipstick and beard are the most sensitive attributes.
Similarly, we discovered that changing specific attributes can largely affect the outcome of a well-trained keypoint detection model.  Interestingly, both keypoint detectors are very sensitive to mustache and eyeglasses, and not very sensitive to hair color or perceived gender. This is expected,  since keypoints have a higher density around the eyes and mouth region, and modification of these regions can be critical to the accuracy. 

Fig.~\ref{fig:demo_images} shows example images of \ourmodel attacking the two target models. For the perceived-gender classifier in Fig.~\ref{fig:demo_images} (a), we can see from the first four columns that mutating the lipstick and beard attributes will influence the model's prediction. The last three columns show that mutating other attributes including hair color, skin color, and bangs can also affect the model decision. Fig.~\ref{fig:demo_images} (b) shows that \ourmodel changes attribute such as eyeglasses, pale skin, or mustache to cause keypoints misdetection in facial images. This sensitivity analysis and adversarial examples can provide insights into the kind of images where the keypoint detector or classifier fails, and generate adversaries to improve performance. More adversarial examples and histograms for the remaining attributes are shown in Appendix A and B.

\begin{table*}[h]
   \centering
   \footnotesize
      \begin{tabular}{@{}lcccccccccccc@{}}
     \toprule
     & \multicolumn{6}{c}{PSNR ($\uparrow$)}     & \multicolumn{6}{c}{SSIM ($\uparrow$)}  \\
     \cmidrule(lr){2-7} \cmidrule(lr){8-13}
     & SPT-50   & \makecell{\ourmodel-50\\(Attr)} & \makecell{\ourmodel-50\\(Full)}  & SPT-200    & \makecell{\ourmodel-200\\(Attr)}   & \makecell{\ourmodel-200\\(Full)}   & SPT-50   & \makecell{\ourmodel-50\\(Attr)} & \makecell{\ourmodel-50\\(Full)}  & SPT-200    & \makecell{\ourmodel-200\\(Attr)}   & \makecell{\ourmodel-200\\(Full)}   \\
     \midrule
     $\displaystyle \eta = \frac{0.25}{255}$  &26.63 &41.98 & \textbf{42.24} &19.43  &31.21 & \textbf{33.94}  &0.9083  &0.9929 & \textbf{0.9930} &0.7732  &0.9602 & \textbf{0.9718}  \\
     \midrule
     $\displaystyle \eta = \frac{4}{255}$  &14.18   &22.36& \textbf{28.25} &13.59 &20.16 & \textbf{25.75}   &0.6385   &0.8573 & \textbf{0.9285} &0.6230 &0.8037 & \textbf{0.8926}   \\
     \bottomrule
   \end{tabular}
      \caption{\label{tab:image_quality_evaluation} Image quality evaluation for \ourmodel and SPT.}
\end{table*}

\vspace{-3mm}
\subsubsection{Image quality evaluation }
We evaluated the image perceptual quality for adversarial examples generated by SPT \cite{2019semanticadversarial} and \ourmodel. To interpret Table \ref{tab:image_quality_evaluation}, SPT-50 ($\eta = \frac{0.25}{255}$) stands for the adversarial examples generated by SPT with 200 iterations and step size of $\frac{0.25}{255}$. The tables show that SIA's image quality is better than SPT under both PSNR (Peak Signal to Noise Ratio) and SSIM (Structured Similarity Indexing Method) \cite{SSIM} metrics.
We can see that perturbing in both image space and attribute space produces visually finer adversarial images. In fact, SIA’s  pixel-level perturbation helps to avoid exaggerated semantic variation that makes the image generation collapse.

\subsubsection{Sensitivity by single-attribute optimization}

\begin{figure}[t]
    \centering
    \includegraphics[width=\linewidth]{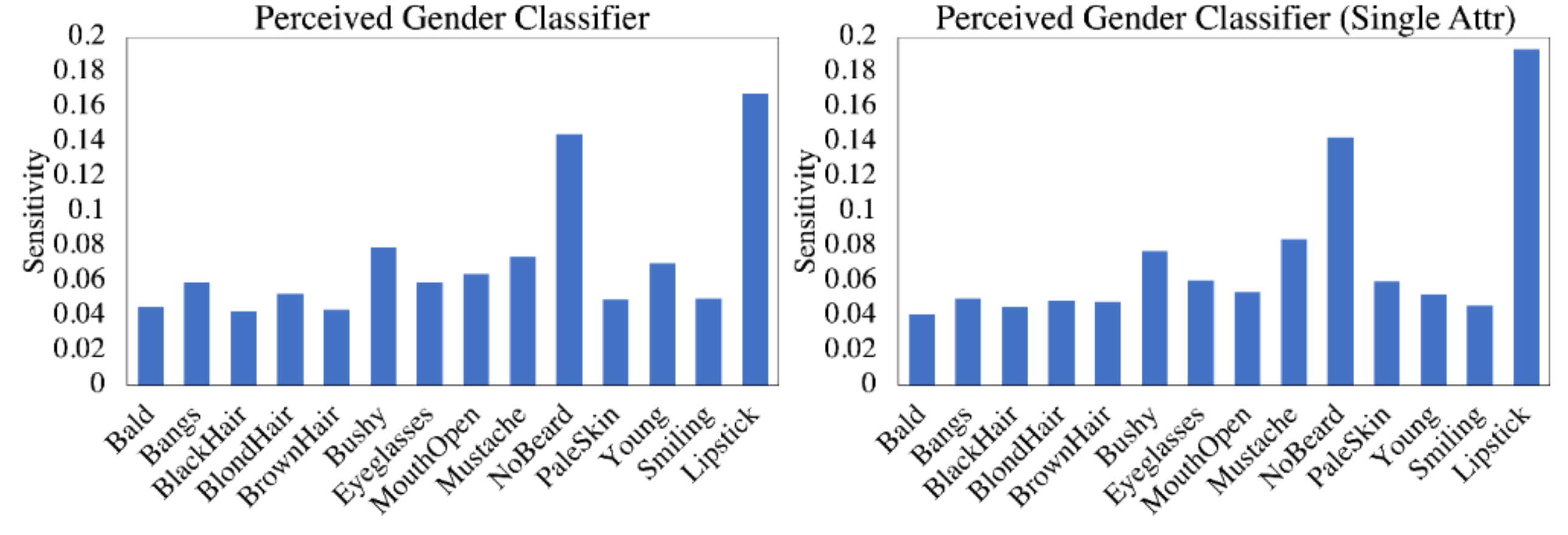}
    \caption{Histogram for attribute sensitivities (under multi- and single-attribute optimization) for the perceived-gender classifier.}
    \label{fig:single_attr_hist}
    \vspace{-3mm}
\end{figure}

We can also perform \ourmodel independently for every single attribute and organize the sensitivities as the histogram on the right in Fig.~\ref{fig:single_attr_hist}. We can see that \ourmodel's histograms, no matter multi-attribute or single-attribute, support consistent analysis of most sensitive attributes. However, it is worth noting that a greedy single attribute perturbation can be computationally expensive for a large attribute space (e.g., 15 attributes). It is very time-consuming to adversarially traverse a single attribute over the dataset and repeat 15 times (i.e., repeat for each attribute) in a grid-search manner. Jointly optimizing all attributes is more time-effective and comprehensive (i.e., exploring a continuous space across all attributes) as the histogram on the left.

\subsection{Attack Effectiveness} \label{attack_effectiveness}

This section compares \ourmodel to popular gradient-based adversarial attacks in a white-box setting for the attribute classifiers. Then an ablation study is conducted to demonstrate the effectiveness of various components in \ourmodel, including the use of attribute and image perturbations. 
\subsubsection{Attack success rate}
\ourmodel constrains the perturbation bounds of attribute space ($\mathbf{a}$) and image space ($\mathbf{x}$) separately. The attributes that do not overlap with the target model range between $[0,1]$ with a step size $\eta = \frac{0.25}{255}$. The attribute that is equivalent to the target classifier is constrained to be a small constant depending on the attribute being classified. 
We iteratively perturbed the input image bounded by $\epsilon = \frac{1.5}{255}$ with $\eta = \frac{0.25}{255}$. The number of steps for both the attribute space and image space will be $200$.  The evaluated subset in CelebA corresponds to the first $10,000$ images. 

We used the FGSM \cite{GoodfellowSS14} and PGD \cite{madry2018towards} under $l_{\infty}$ norm of different perturbation bounds as baseline methods. For PGD, the iteration step size $\eta = \frac{0.25}{255}$ with 200 steps in total. For FGSM, the attack will iterate once within a bounded perturbation. In the adversarial training experiment, we additionally compared with SPT \cite{2019semanticadversarial} using the same attribute space as \ourmodel. However, we did not compare with \cite{Li_2021_CVPR} since their method samples adversarial examples from StyleGAN, which does not support attacking existing images.  We also did not compare with \cite{Lang_2021_ICCV} because their method requires training a separate model on StyleGAN's original training data for each target model.

 Table~\ref{tab:successful_rate} shows the attack success rates for different perturbation-based attacks on multiple target classifiers. 
Notably, we can see that \ourmodel achieves performance comparable to traditional attacks with smaller perturbations.

\begin{table}[t]
   \centering
   \footnotesize
      \begin{tabular}{@{}lcccc@{}}
     \toprule
    \multicolumn{1}{c}{$\epsilon$} & Classifier &FGSM    & PGD    & \ourmodel   \\
     \midrule
     \multicolumn{1}{c}{\multirow{2}{*}{1.5/255}} & Eyeglasses &28.01 & 49.85 & \textbf{68.20} \\
                        & Perceived Gender & 56.31& 65.32& \textbf{88.19}  \\
                        \midrule
     \multicolumn{1}{c}{\multirow{2}{*}{2/255}} & Eyeglasses &42.83 & 87.79 & \textbf{94.82} \\
                        & Perceived Gender & 75.27 & 87.19 & \textbf{92.44}  \\
                        \midrule
     \multicolumn{1}{c}{\multirow{2}{*}{4/255}} & Eyeglasses &78.90 & 99.94& \textbf{99.99} \\
                        & Perceived Gender & 97.33 & 97.41& \textbf{98.53}  \\
     \bottomrule
   \end{tabular}
      \caption{ \label{tab:successful_rate} Success rate (\%) for different adversarial attack methods with different perturbation bounds.}
   \vspace{-3mm}
\end{table}

\subsubsection{Ablation for image v.s. attribute space} This experiment analyzes the attack success rate when attacking the image and/or the attribute space (see Table~\ref{tab:component_study}). Original reconstruction refers to the images reconstructed by $\mathcal{G}_\phi$ without any perturbations. I/A refers to only updating the image/attribute space during the attack. PI/PA refers to partially updating the image/attribute space in the first $20$ iterations of the total $200$ iterations. 
Note that the Attr-space setting is different from SPT~\cite{2019semanticadversarial} since \ourmodel uses $\operatorname{sign}$ linearization to constrain the gradient updates to stabilize the attack. 
As expected, the attack effectiveness is much higher regardless of using attribute space alone or in combination.  

\begin{table}[t]
   \footnotesize
   \centering
   \begin{tabular}{@{}lcccc@{}} 
     \toprule
     & \multicolumn{1}{c}{Eyeglasses} &  \makecell{Goatee} & \multicolumn{1}{c}{Age} & \multicolumn{1}{c}{Sideburns}     \\
     \midrule
    \multicolumn{1}{c}{OR}                & 0.02  & 3.49  & 0.15  & 3.32\\
    \multicolumn{1}{c}{I}           & 1.83  & 13.58 & 6.10  & 12.51\\
    \multicolumn{1}{c}{I + PA}     & 18.25 & 26.54 & 30.16 & 22.46\\
    \multicolumn{1}{c}{A}            & 32.67 & 83.59 & \textbf{90.90} & 67.97\\
    \multicolumn{1}{c}{A + PI}    & 44.26 & 79.82 & 85.98 & 70.65\\
    \multicolumn{1}{c}{Full}                  & \textbf{68.20} & \textbf{90.51} & 90.38 & \textbf{87.08}\\
     \bottomrule
   \end{tabular}
   \caption{\label{tab:component_study} Ablation study on the success rate (\%) of attacks. }
    \vspace{-3mm}
\end{table}

\subsubsection{Extending attribute space} 
\begin{figure}[t]
    \centering
    \begin{subfigure}[b]{0.49\linewidth}
    \includegraphics[width=\linewidth]{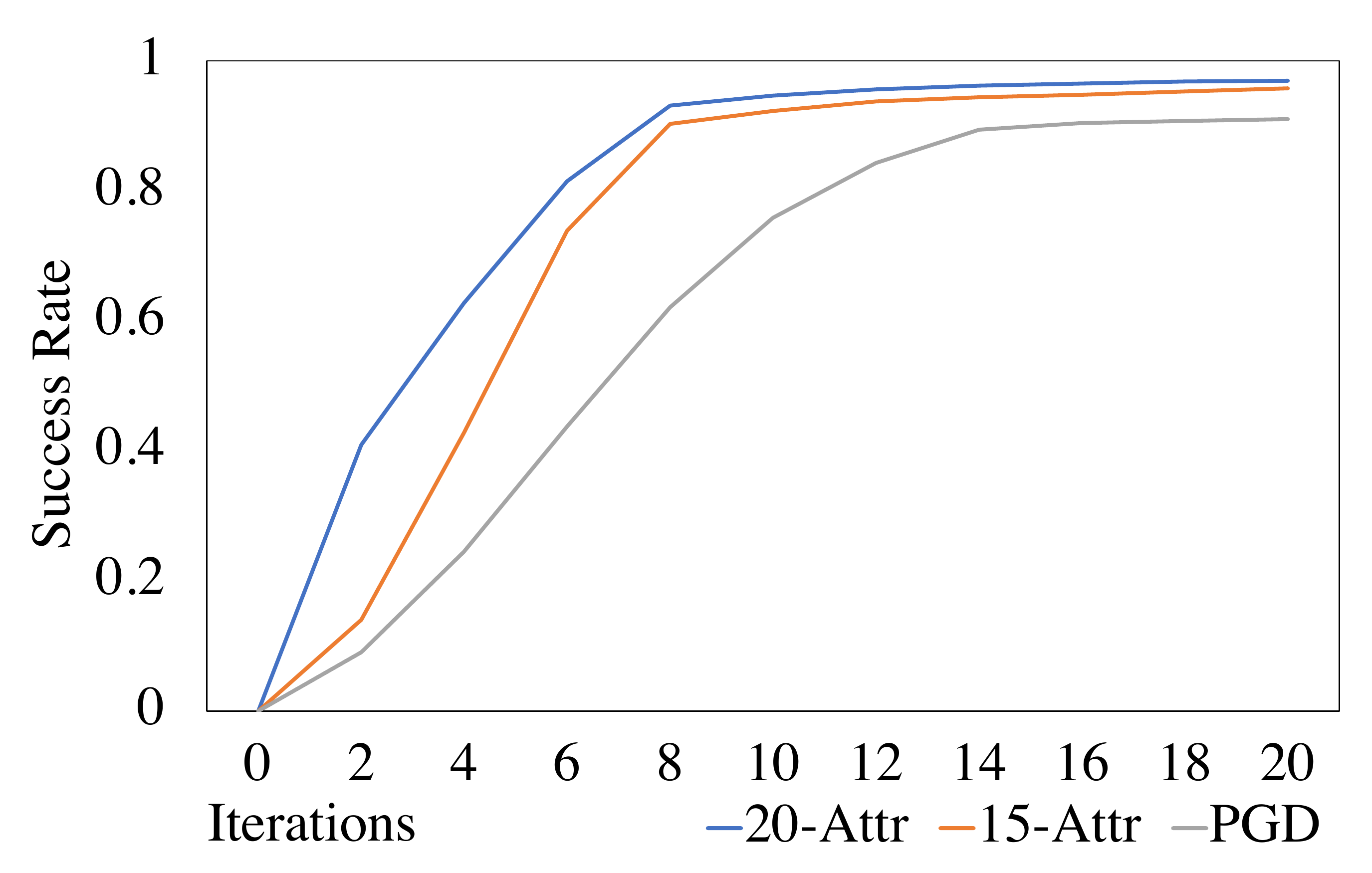}
    \caption{Ablation for attack iterations}
    \end{subfigure}
    \centering
    \begin{subfigure}[b]{0.49\linewidth}
    \includegraphics[width=\linewidth]{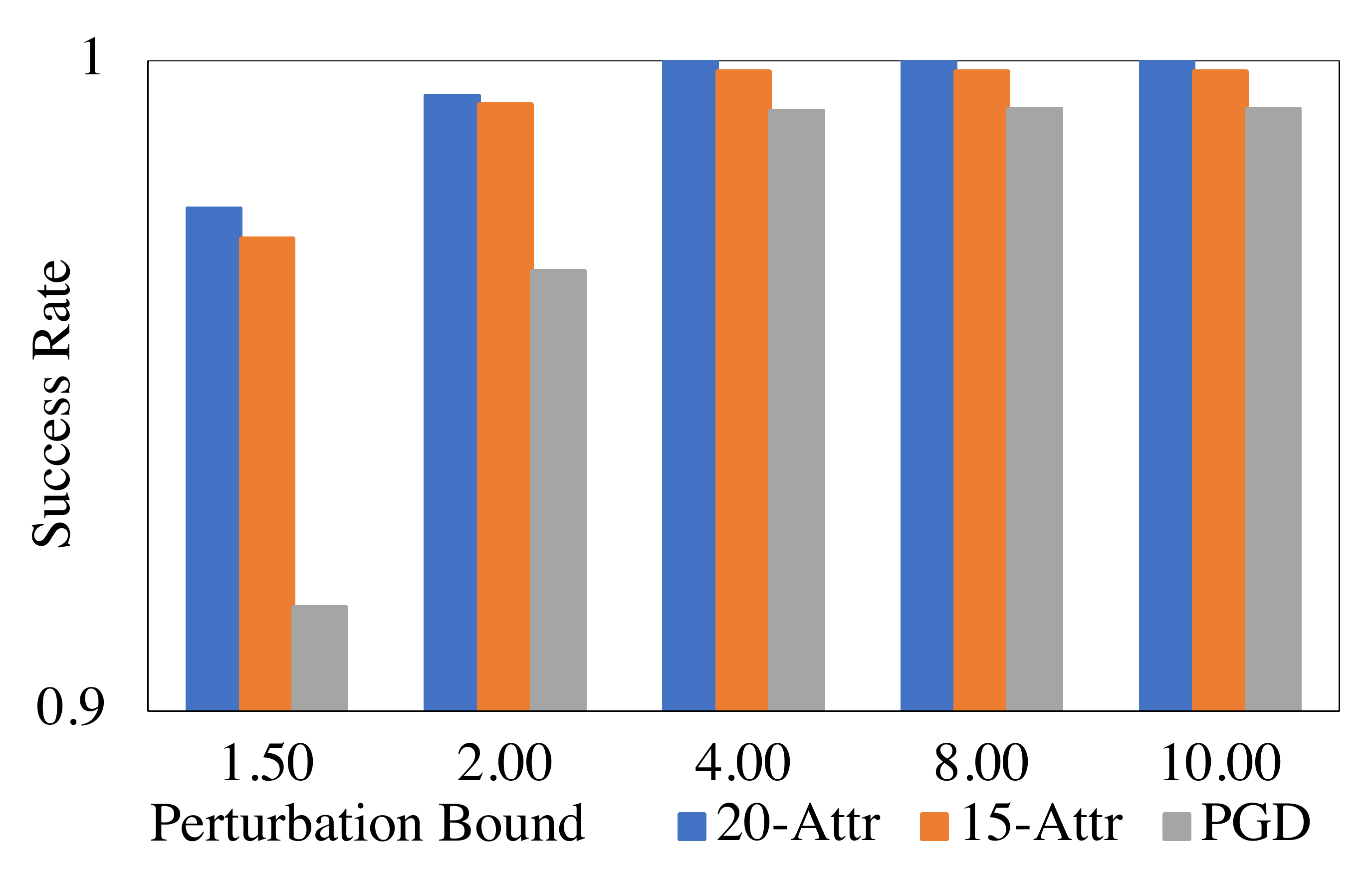}
    \caption{Ablation for attack bound}
    \end{subfigure}
    \caption{The effect of different attribute spaces used in \ourmodel. We compare $15$ and $20$ attributes and show that a larger space of attributes leads to faster attack convergence (left) and a higher success rate with various bounds (right).     }
    \label{fig:ablation_20_attr}
    \vspace{-1mm}
\end{figure}

We experimented with an alternative attribute space of 
$20$ attributes for $\mathcal{G}_\phi$. We removed Black\_Hair, Brown\_Hair, Bushy\_Eyebrows, Eyeglasses, Male(perceived), No\_Beard, Young(perceived), Wearing\_Lipstick which are either attribute of target classifier or attributes with overlapped concepts. Then we added Narrow\_Eyes, Oval\_Face, Pale\_Skin, Pointy\_Nose, Receding\_Hairline, Rosy\_Cheeks, Sideburns, Straight\_Hair, Big\_Lips, Big\_Nose, Chubby, Goatee, Heavy\_Makeup, High\_Cheekbones.  Compared with the attribute space used in our main experiment, this alternative attribute space covers more semantic variations in facial data.  Fig.~\ref{fig:ablation_20_attr}(a) shows the success rate for the attractive classifier. PGD refers to the implemented PGD attack~\cite{madry2018towards}. The larger the attribute space, the higher the success rate, and the attack converges in fewer iterations.  
This is not surprising because the larger semantic space helps $\mathcal{G}_\phi$ to search the combination of adversarial attributes more effectively. Fig.~\ref{fig:ablation_20_attr}(b) shows that with the same perturbation bound setting, the extended $\mathcal{G}_\phi$ will give a stronger attack on the target classifier. 

\subsection{Adversarial Training} \label{adversarial_training}

\begin{figure}[t]
    \centering
    \includegraphics[width=\linewidth]{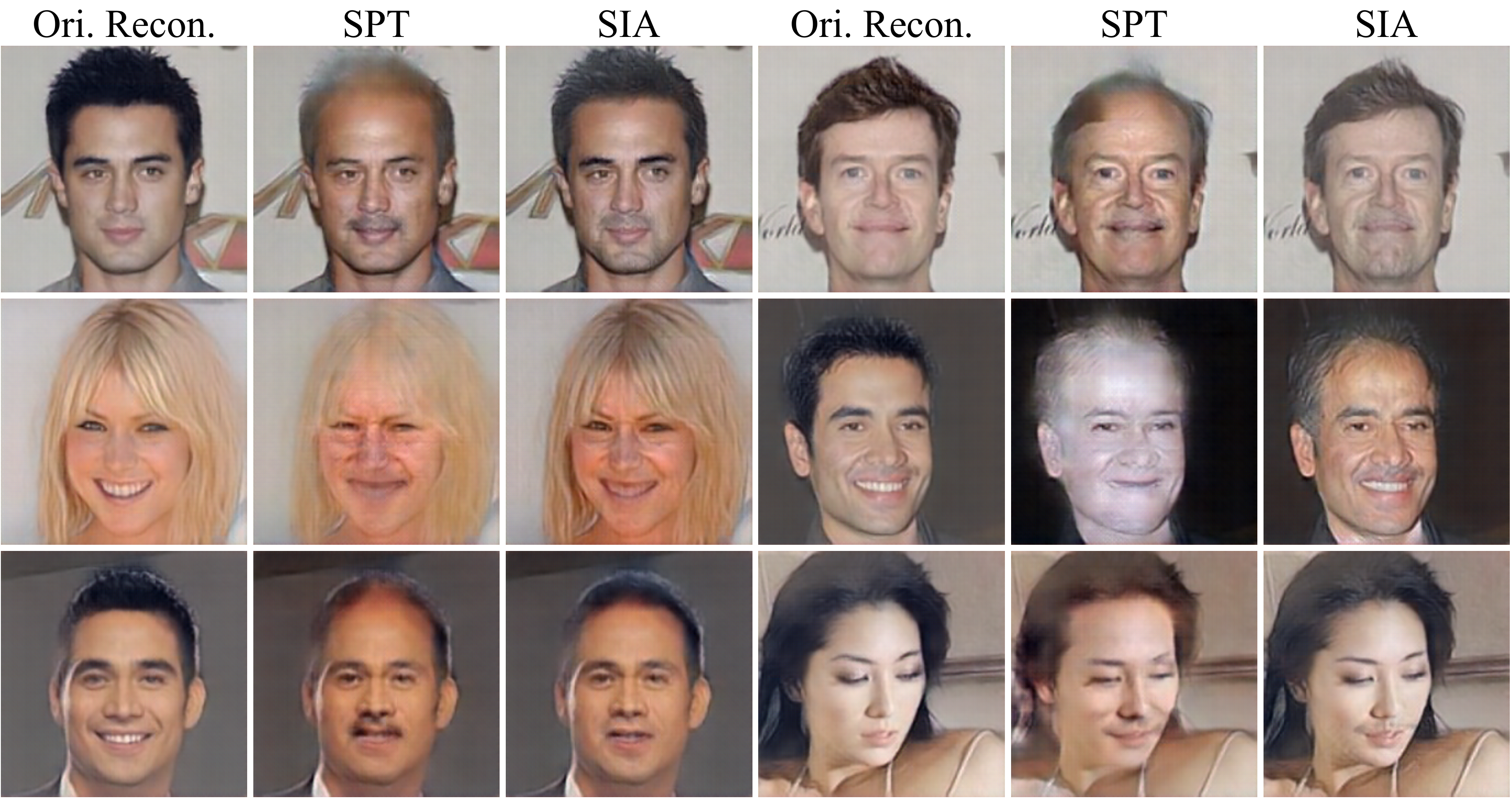}
    \caption{ Demonstration of \ourmodel and SPT \cite{2019semanticadversarial} adversarial examples on the eyeglasses classifier. Results show that SPT generates unrealistic images, while \ourmodel generates realistic images with small but semantic modifications of the original image.
    }
    \label{fig:demo_baselines}
    \vspace{-0mm}
\end{figure}

 In this experiment, we evaluated the effectiveness of SIA to improve adversarial robustness. We adopted
the setting such that the target model is fine-tuned with adversarial examples for one epoch. Table~\ref{tab:robustness} shows how \ourmodel can be used effectively for re-fitting adversarial examples generated by Algorithm 1. \ourmodel-Adv, PGD-Adv, SPT-Adv are eyeglasses classifiers adversarially trained with $30,000$ adversarial examples generated by the corresponding attack method from the first $30,000$ images of CelebA. The perturbation bound is $\epsilon = 1.5/255$. Non-adversarial training means the regular classifier trained in Section~\ref{visual_model_dianosis}. All models are evaluated on the first $10,000$ images from the CelebA test set that the models have never seen before. Results show that the robustness of \ourmodel adversarial training is transferable to other attack methods, but not vice versa (i.e., see how the column SIA-Adv works well across all the attacks). This is because our adversarial example constructs both conceptual shifts in the semantic space and noise shift in the image space, which introduces richer information during the adversarial training compared to traditional perturbation attacks. Fig.~\ref{fig:demo_baselines} shows the visual comparison of \ourmodel and SPT adversaries on the eyeglasses classifier. SPT generates less fine-controlled semantic changes because updating only the attribute space results in large changes across many attributes. More visual comparisons of different baselines for adversarial training are reported in Appendix C.

\begin{table}[t]
  \centering
  \scriptsize
  \begin{tabular}{@{}lccccc@{}}
     \toprule
                                               & Non-Adv    & \makecell{PGD-Adv \cite{madry2018towards}} & SPT-Adv \cite{2019semanticadversarial}  & \ourmodel-Adv \\
     \midrule
     Clean Test Set                                   & 99.63 & 99.54 & 99.51 & 99.52 \\
     \midrule
     FGSM ($1.5/255$)                  & 73.97 & 86.55 & 77.76 & \textbf{94.01} \\
     PGD ($1.5/255$)                   & 50.63 & 81.98 & 16.52 & \textbf{86.90} \\
     \ourmodel ($1.5/255$)             & 12.59 & 27.90 & \textbf{74.01} & 67.07 \\
     FGSM ($4/255$)                    & 22.47 & 22.09 & 41.78 & \textbf{45.51} \\
     \ourmodel ($4/255$)               & 4.56 & 10.85 & 10.55 & \textbf{12.57} \\
     \bottomrule
  \end{tabular}
  \caption{\label{tab:robustness} Adversarial training. The reported numbers represent the accuracy (\%) for adversaries.}
  \vspace{-3mm}
\end{table}

\vspace{-2mm}
\subsubsection{Standard deviation of attribute robustness} We established a measure named Standard Deviation of Attribute Robustness (SDAR) to understand the final variance of our model across attributes.  For a given classifier $f_\theta$, \ourmodel generates the sensitivity histogram based on the attribute perturbation vector $\mathbf{s}$ of length $L$. The SDAR metric $\sigma_{\bm{s}}$ is defined as the standard deviation of the sensitivity values $\sigma_{\bm{s}}=\sqrt{\frac{1}{L} \sum_{i=1}^{L}\left(s_{i}-\bar{\mathbf{s}} \right)^{2}}$.

Ideally, an unbiased model should have equal sensitivity across all attributes, hence a decrease in the standard deviation will indicate that the model is less biased. To validate the method, we calculated SDARs after evaluating different models from adversarial training. The test data was the first $10,000$ images of CelebA test set. We evaluated the SDAR metric under two bounds ($\epsilon$) of \ourmodel. Table \ref{tab:SDAR_appendix} shows the results. We can see that the non-adv classifiers will have larger $\sigma$ and the adversarially-trained models have smaller $\sigma$ since the model generalization is improved by the adversarial training process. By optimizing both the attribute and the image space, \ourmodel-Adv better generalizes over attributes than regular classifiers.

\begin{table}[t]
  \centering
  \footnotesize
  \begin{tabular}{@{}lc@{}c@{}c@{}c@{}}
     \toprule
                                               & Non-Adv \    & \  \makecell{PGD-Adv \cite{madry2018towards}} \ & \ SPT-Adv \cite{2019semanticadversarial} \   & \ \ourmodel-Adv (Ours) \\
     \midrule
     $\displaystyle \epsilon = \frac{1.5}{255}$             & 0.1145 & 0.0917 & 0.0852 & \textbf{0.0781}\\
     \midrule
     $\displaystyle \epsilon = \frac{4}{255}$             & 0.1419 & 0.1270 & 0.1253 & \textbf{0.1056}\\
     \bottomrule
  \end{tabular}
  \caption{\label{tab:SDAR_appendix} SDAR metric on different adversarially-trained models from Section 4.4. A lower value indicates a less biased model.}
\end{table}

\subsection{Robustness to Imbalanced Datasets} \label{data_augmentation_experiment}
This section reports experiments to evaluate the robustness of \ourmodel in learning from imbalanced datasets. In these situations, it is vital to develop algorithms that are not biased toward the majority class. While data augmentation and re-weighting are commonly used techniques, we show how \ourmodel provides an alternative that generates semantically meaningful augmentation with high visual quality.

We trained two attribute classifiers, for eyeglasses and bangs, using the ResNet50 architecture. We generated a synthetically unbalanced dataset to produce a controlled imbalanced environment. For training, we randomly sampled $30,000$ images from CelebA training set such that $1\%$ are positive and $99\%$ are negative. We trained the classifiers from random initialization. For testing, we use balanced test data including random $2,500$ positive-label images and $2,500$ negative-label images from CelebA test set.

Table \ref{tab:data_augmentation} shows the precision, recall, and accuracy for several imbalanced learning strategies. We compared \ourmodel to five data augmentation and re-balancing approaches. The Non-Adv attribute classifiers are trained on the synthetic data with 1:99 CelebA training set.
The CutMix \cite{2019cutmix} baselines are augmented with action probability $p = 0.5$ and learning rate $\alpha = 0.001$. We followed PyTorch's implementation on AutoAugment \cite{Cubuk_2019_CVPR_autoaugment}.
We also included two commonly used baselines to deal with imbalanced data: Reweighting and Resampling. 
Reweighting means upweighting the under-represented samples based on the proportion of class samples. 
Resampling means duplicating the under-represented samples until different classes have the same number of samples. 
\ourmodel refers to classifiers augmented by randomly sampling $30,000$ our adversarial images. \ourmodel + Reweight is the scheme where the reweighting is performed on our \ourmodel-augmented dataset. Results show that \ourmodel can effectively be used to augment imbalanced datasets, outperforming other widely used augmentation methods. One possible reason is that \ourmodel generates semantically meaningful augmentations, different from CutMix and AutoAugment. Finally, we conduct a similar experiment with pre-trained classifiers in Appendix D. We show that the difference in accuracy between the methods narrows down considerably if we pre-train the classifiers. This is not surprising, since pre-training with sufficient data provides robust features that are less prone to imbalance.
\begin{table}[t]
   \centering
   \footnotesize
   \begin{tabular}{@{}b{0.02\textwidth}lccc@{}} 
     \toprule
      & Strategy & Prec. $\uparrow$ & Recall $\uparrow$ & Acc. (\%) $\uparrow$ \\
     \midrule
     \multirow{7}{*}{\rotatebox[origin=c]{90}{Eyeglasses}} & Non-adv classifier              & 0.9985 & 0.8052 & 90.20  \\
     & Reweighting                     & \textbf{0.9995} & 0.8368 & 91.82 \\ 
     & Resample                        & 0.9984 & 0.7700 & 88.44 \\
     & CutMix                          & 0.9963 & 0.3236 & 66.12 \\
     & AutoAugment                     & 0.9975 & 0.8004 & 89.92 \\
         & \ourmodel-Adv (ours)                & 0.9991 & \textbf{0.8864} & \textbf{94.28} \\
     & \ourmodel-Adv + Reweight (ours)     & 0.9991 & 0.8856 & 94.24 \\
     \midrule
             \multirow{7}{*}{\rotatebox[origin=c]{90}{Bangs}} & Non-adv classifier              & 0.9847 & 0.2576 & 62.68 \\
     & Reweighting                     & 0.9912 & 0.2708 & 63.42 \\ 
     & Resample                        & 0.9935 & 0.1840 & 59.14 \\
     & CutMix                          & \textbf{1.0000} & 0.0000 & 50.00 \\
     & AutoAugment                     & 0.9701 & 0.0260 & 51.26 \\
         & \ourmodel-Adv (ours)                & 0.9791 & 0.4116 & 70.14 \\
     & \ourmodel-Adv + Reweight (ours)     & 0.9854 & \textbf{0.5960} & \textbf{79.36}\\
     \bottomrule
   \end{tabular}
   \caption{\label{tab:data_augmentation} Comparison of different strategies for learning from imbalanced datasets. See text.}
   \vspace{-0mm}
\end{table}

\section{Conclusions and Future Work} \label{conclusion_and_future}

This paper introduced \ourmodel, an attribute-assisted adversarial method with applications in model diagnosis, improving target model robustness, and increasing the success of visual attacks. A major appeal of our technique is the capacity of analyzing a deep learning model without
a carefully designed test set.
\ourmodel reveals the dependencies between attributes and model outputs, which helps interpret the biases learned by models during prediction. We hope our results pave the way for new tools to analyze models and inspire future work on mitigating such biases.

While we showed the benefits of our technique in two computer vision problems, our approach is applicable to any end-to-end differentiable target deep learning model. It is unclear how to extend this approach to non-differentiable ML models, and more research needs to be done. Our method works with white-box attacks since our primary motivation is to diagnose a known model. More research needs to be done to address black-box attacks. Furthermore, we have illustrated the power of \ourmodel only in the context of faces, but our method can extend to generative models that have been trained with other attributes of interest and can be applied to other visual domains.

{
\small
\bibliographystyle{ieee_fullname}
\bibliography{arxiv}
}
\newpage
{\onecolumn 
\appendix
\section*{Supplementary Material}
\section{\ourmodel Adversarial Images}
Fig. \ref{fig:demo_images_appendix_1} and Fig. \ref{fig:demo_detector_appendix_Fakeit} show additional examples of adversarial images generated by \ourmodel for different target models. We find the adversarial semantic shifts on keypoint detectors are visually greater than that on classifiers. These results show that detectors are more robust to semantic shifts than classifiers.

\begin{figure*}[!h]
    \centering
    \includegraphics[width=0.75\textwidth]{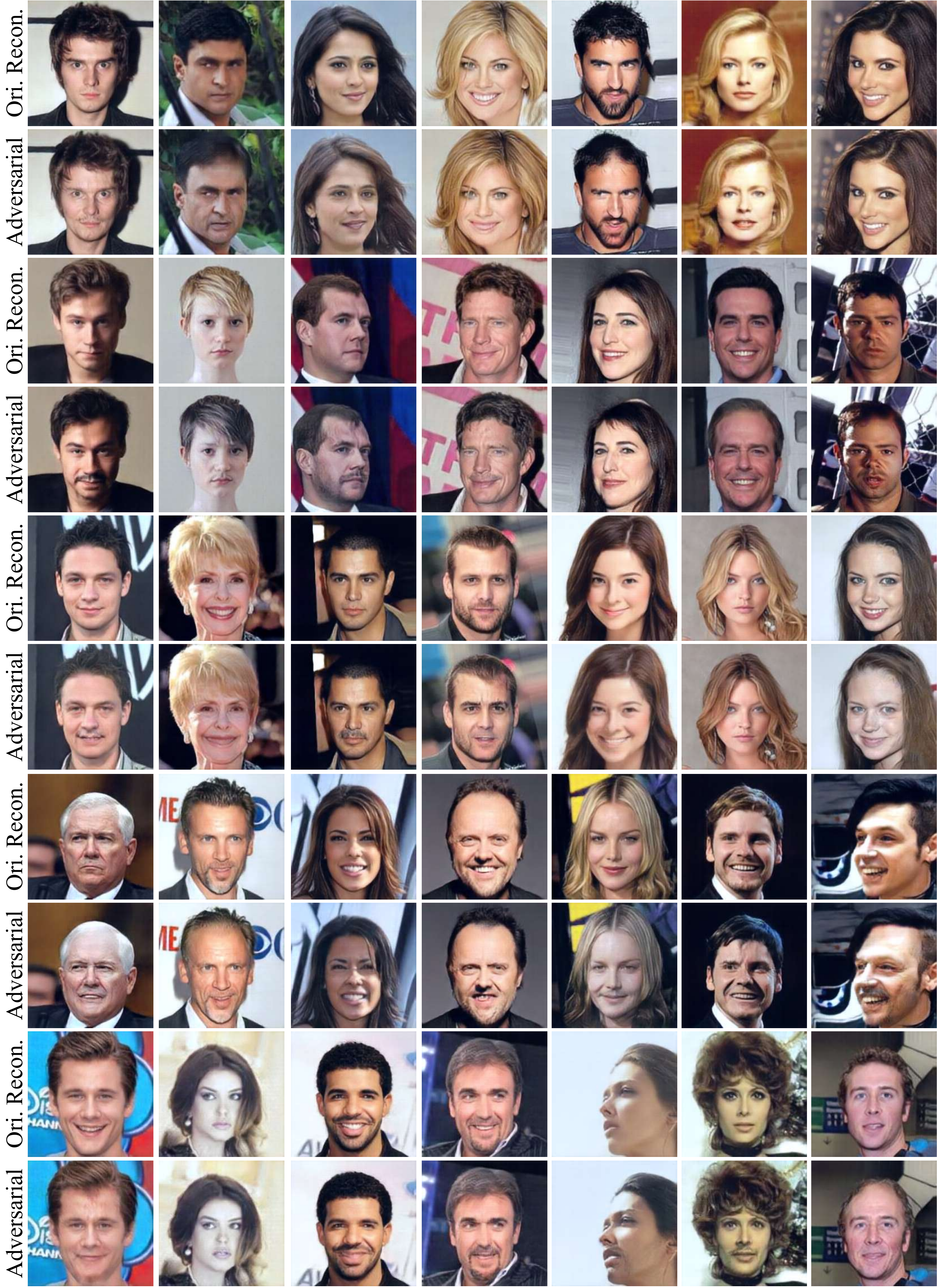}
    \caption{\ourmodel adversarial examples for the eyeglasses classifier.}
    \label{fig:demo_images_appendix_1}
\end{figure*}

\newpage

\begin{figure*}[!h]
    \centering
    \vspace{-5mm}
    \includegraphics[width=0.75\textwidth]{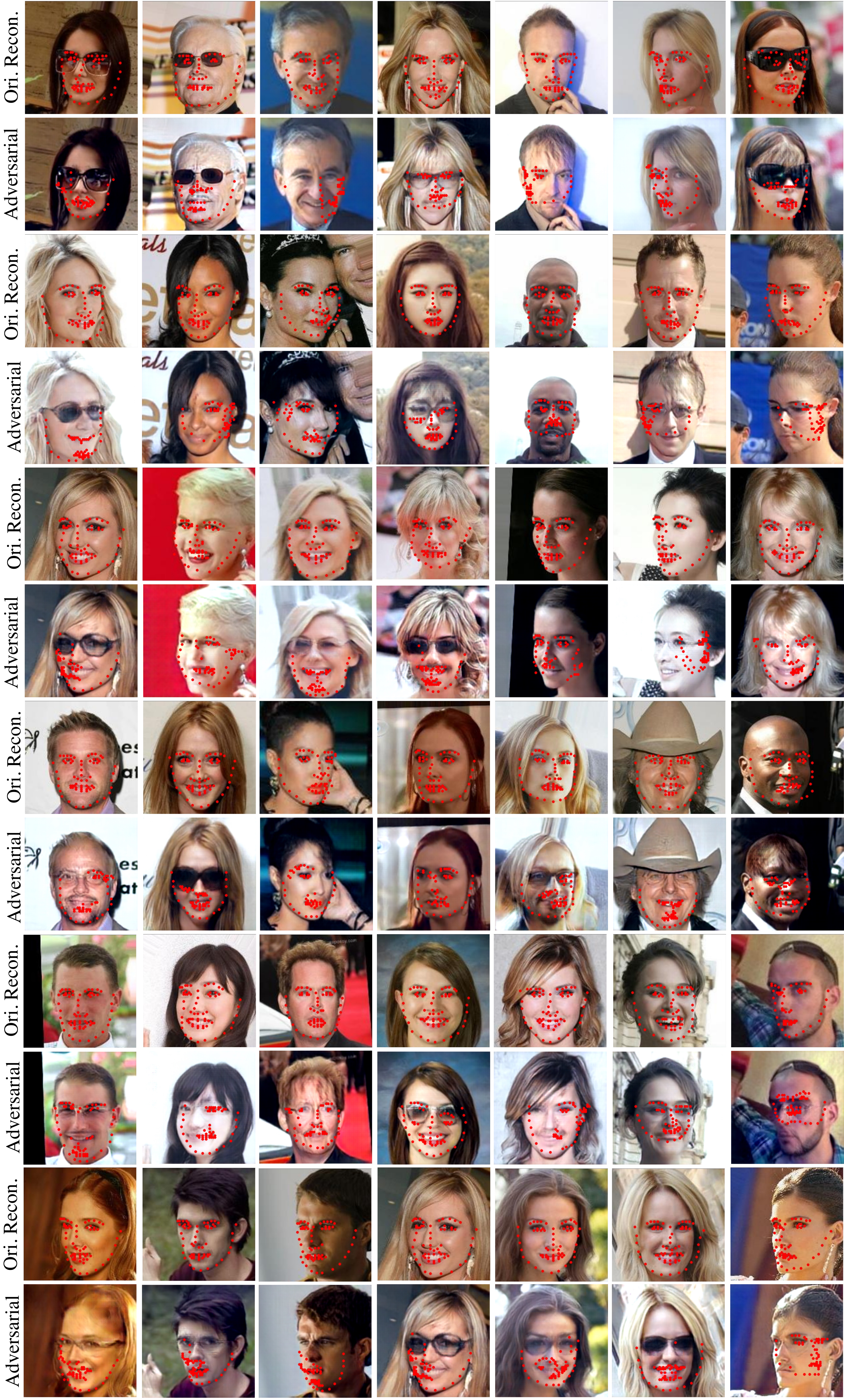}
    \caption{\ourmodel adversarial examples for the keypoint detector trained on Microsoft Fake-it \cite{wood2021fake}.}
    \label{fig:demo_detector_appendix_Fakeit}
\end{figure*}

\newpage
\section{Histograms for Model Interpretation }

Fig.~\ref{fig:demo_histogram_appendix} shows the histograms of the sensitivity across attributes generated for additional  attribute classifiers in Section 4.2. 

\begin{figure*}[!h]
    \centering
    \includegraphics[width=0.75\textwidth]{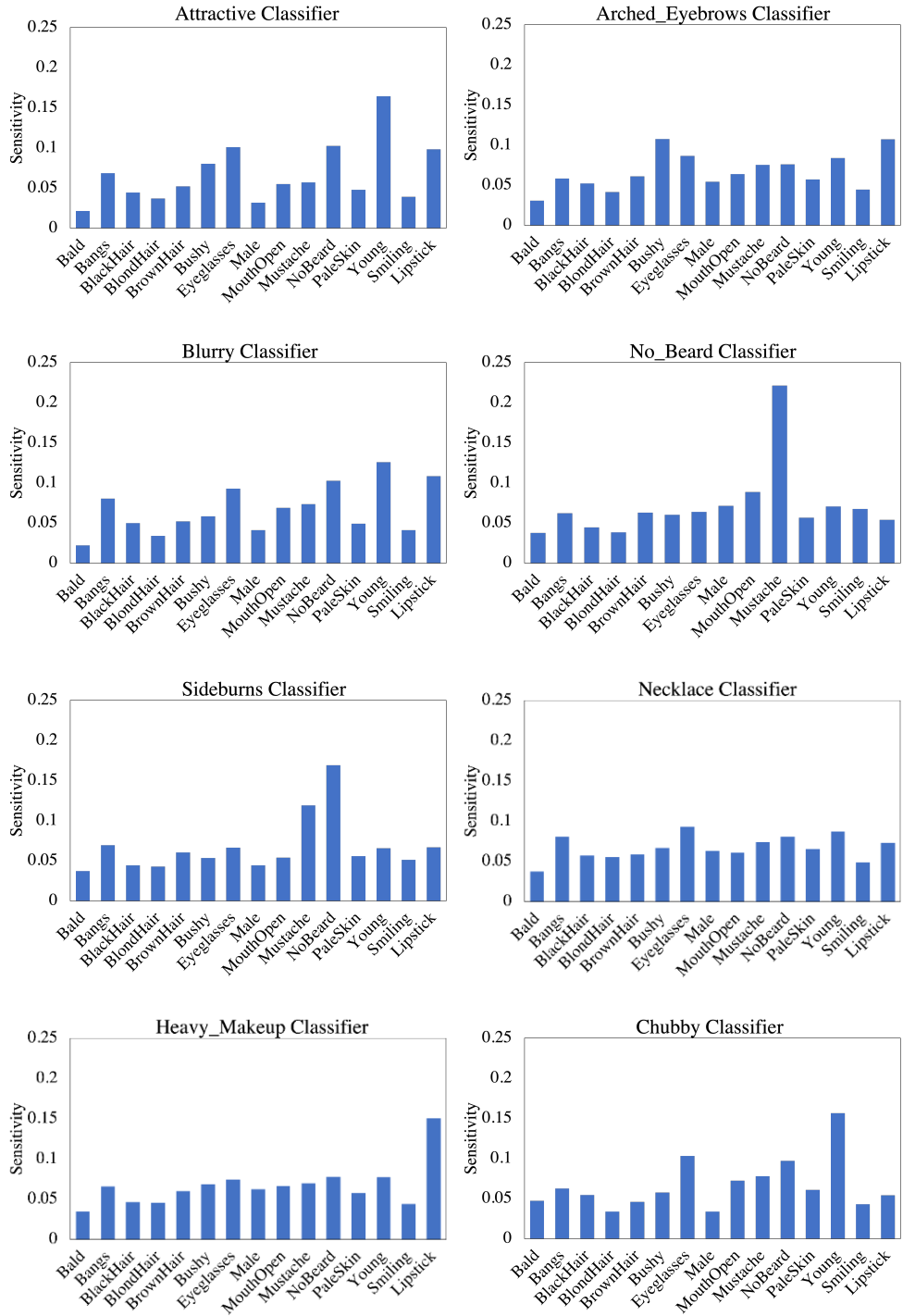}
    \caption{Attribute sensitivity analysis generated by \ourmodel for more target classifiers.}
    \label{fig:demo_histogram_appendix}
\end{figure*}

\section{Image Synthesis Analysis}
\subsection{Visual Comparison of Adversarial Examples}
Fig. \ref{fig:demo_visual_comparison} shows more visual comparisons of the adversarial examples generated by different methods. 
As we can see, \ourmodel adds perturbation in the image space and the attribute space, generating photo-realistic fine-grained adversarial examples. Perturbing in both image space and attribute space produces finer visual adversarial images. In fact, \ourmodel’s pixel-level perturbation helps to avoid exaggerated semantic variation that makes the image generation collapse.

\begin{figure*}[!h]
\centering
\begin{subfigure}[b]{0.1\textwidth}
    \centering
    \caption*{Ori. Recon.}
    \includegraphics[width=\textwidth]{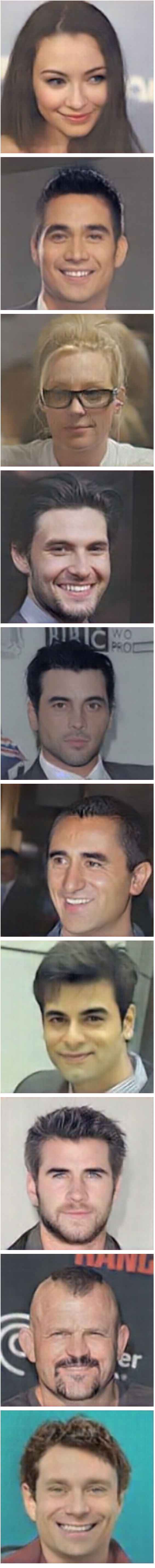}
\end{subfigure}
\hspace{-1.7mm}
\begin{subfigure}[b]{0.1\textwidth}
    \centering
    \caption*{FGSM \cite{GoodfellowSS14}}
    \includegraphics[width=\textwidth]{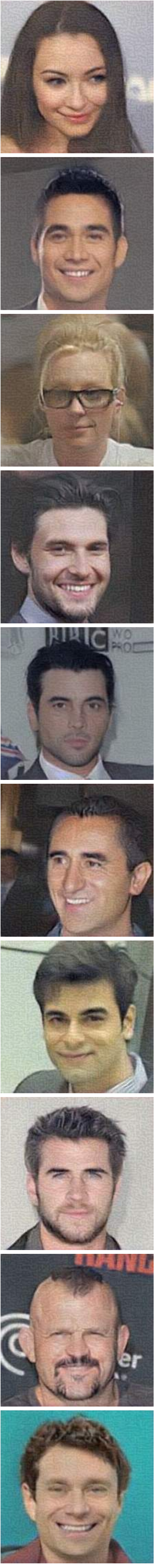}
\end{subfigure}
\hspace{-1.7mm}
\begin{subfigure}[b]{0.1\textwidth}
    \centering
    \caption*{SPT \cite{2019semanticadversarial}}
    \includegraphics[width=\textwidth]{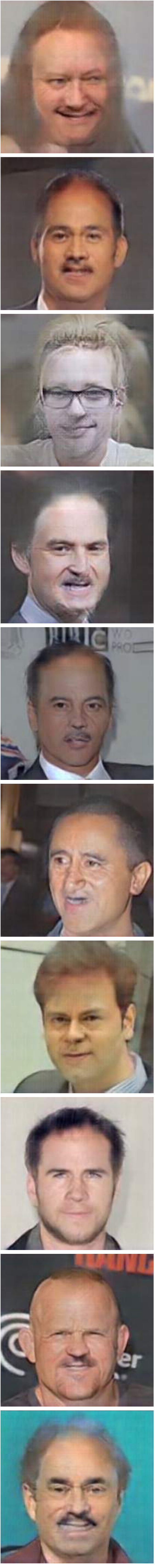}
\end{subfigure}
\hspace{-1.7mm}
\begin{subfigure}[b]{0.1\textwidth}
    \centering
    \caption*{SIA (Ours)}
    \includegraphics[width=\textwidth]{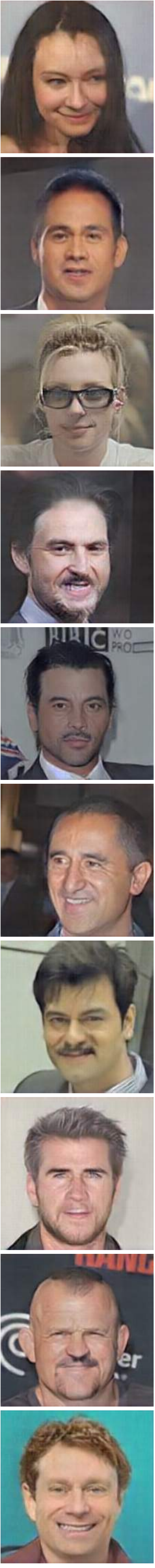}
\end{subfigure}
\hspace{-1.7mm}
\begin{subfigure}[b]{0.1\textwidth}
    \centering
    \caption*{Ori. Recon.}
    \includegraphics[width=\textwidth]{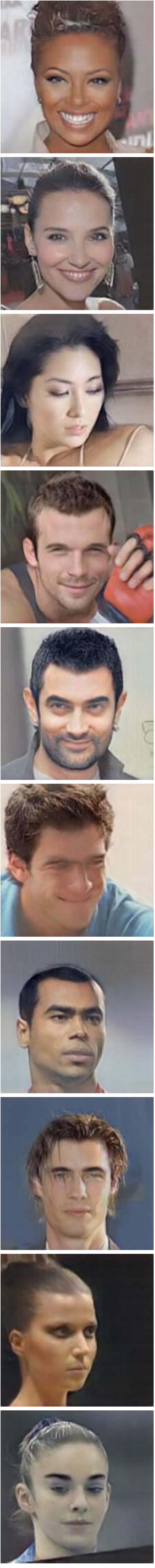}
\end{subfigure}
\hspace{-1.7mm}
\begin{subfigure}[b]{0.1\textwidth}
    \centering
    \caption*{FGSM \cite{GoodfellowSS14}} 
    \includegraphics[width=\textwidth]{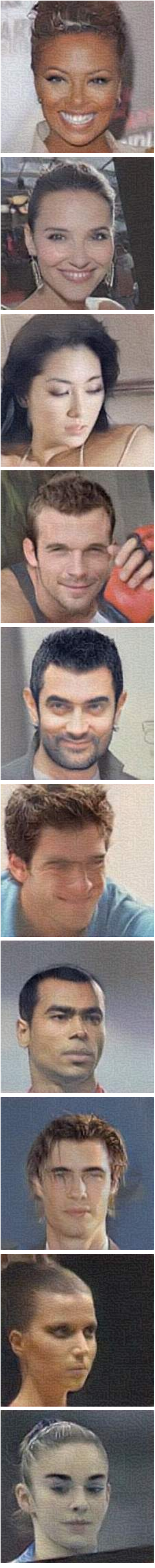}
\end{subfigure}
\hspace{-1.7mm}
\begin{subfigure}[b]{0.1\textwidth}
    \centering
    \caption*{SPT \cite{2019semanticadversarial}}
    \includegraphics[width=\textwidth]{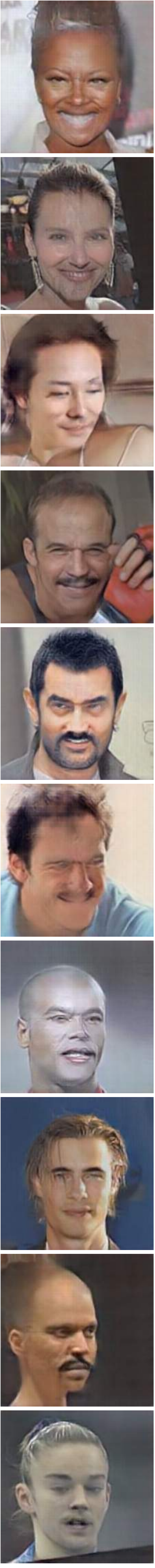}
\end{subfigure}
\hspace{-1.7mm}
\begin{subfigure}[b]{0.1\textwidth}
    \centering
    \caption*{SIA (Ours)}
    \includegraphics[width=\textwidth]{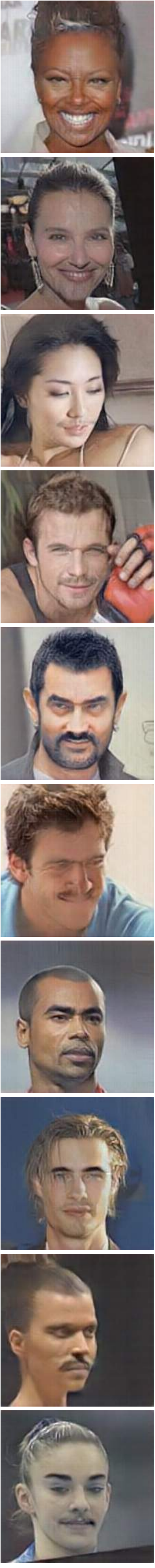}
\end{subfigure}

    \caption{Adversarial examples by FGSM \cite{GoodfellowSS14}, SPT \cite{2019semanticadversarial}, and SIA (Ours) on the eyeglasses classifier. }
    \label{fig:demo_visual_comparison}
\end{figure*}

\subsection{AttGAN's Reconstruction and Semantic Editing}
AttGAN ($\mathcal{G}_\phi$) is capable of editing both fine-level semantics (e.g., beard) and complex concepts (e.g., age). The reconstruction loss during the training of $\mathcal{G}_\phi$ guarantees the preservation of facial details. As stated in \cite{2019attgan}, the use of shortcut layers \cite{2015unet} improves the quality of image translation. During the SPT and \ourmodel attack, we constrained all mutated attributes in the range of [0,1] to make sure that the transformed attribute vector for $\mathcal{G}_\phi$ is valid. The style intensity hyper-parameter is set to $1$, and the number of encoder layers and decoder layers are both $5$.

Additionally, Fig. \ref{fig:appendix_attgan} illustrates the non-adv image projection by our AttGAN backbone $\mathcal{G}_\phi$. We can see from the images that the inverse (flipping) attribute manipulation for most attributes is visually correct. Note that there are naturally hard cases (e.g., modifying images of people with long hair to be bold), and this can be potentially improved in future work.

\begin{figure*}[!h]
    \centering
    \includegraphics[width=\textwidth]{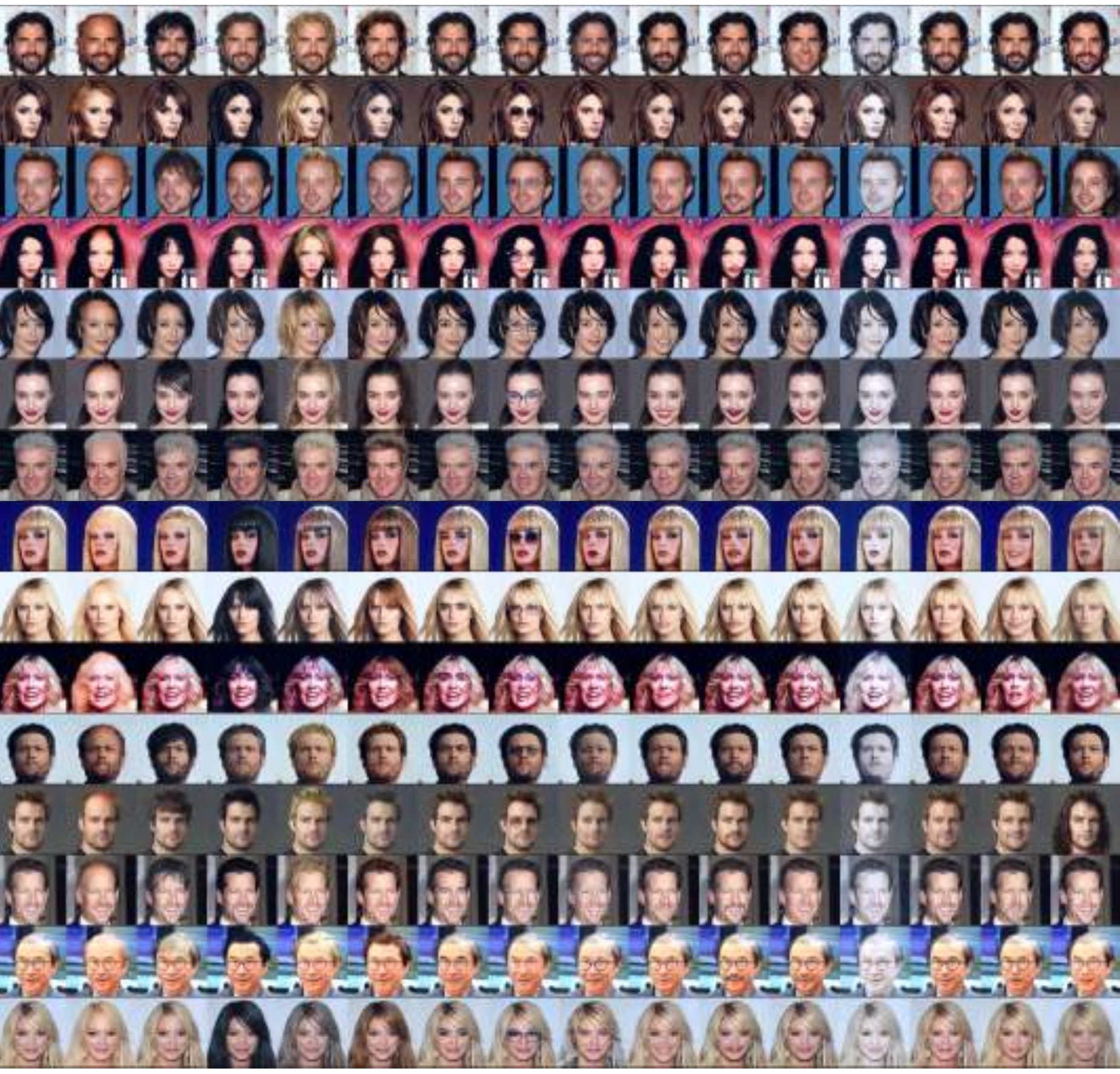}
    \caption{\label{fig:appendix_attgan} $\mathcal{G}_\phi$'s reconstruction and attribute manipulation. The columns are \textit{Reconstruction, Bald, Bangs, Black\_Hair, Blond\_Hair, Brown\_Hair, Bushy\_Eyebrows, Eyeglasses, Gender, Mouth\_Slightly\_Open, Mustache, No\_Beard, Pale\_Skin, Age, Smiling, Wearing\_Lipstick}.}
\end{figure*}

\section{Imbalanced Datasets}

In this section, we ran data augmentation experiments using pre-trained models on a ResNet50 architecture. The pre-trained models are described in Section 4.1. Recall that the CelebA is highly imbalanced for some classes. For instance, the CelebA training set in the case of eyeglasses has $6.46\%$ Positive and $93.54\%$ Negative samples. For the bangs classifier, the imbalance in training data is $15.17\%$ Positive and $84.83\%$ Negative. As in Section 4.5, we used a balanced test data including random $2500$ positive-label images and $2500$ negative-label images from the CelebA test set. Table \ref{tab:data_augmentation_appendix} shows the augmentation results on these two attributes. While \ourmodel still beats conventional techniques in this scenario, it does so by a significantly lesser margin. Fig. \ref{fig:celeba_appendix_full} summarizes the label proportion of each attribute in the CelebA dataset.  As we can notice, there are many attributes that are significantly unbalanced, and therefore the challenge of generating a balanced dataset across attributes.

\begin{table*}[h]
  \centering
  \begin{tabular}{lcccc} 
     \toprule
     & \multicolumn{2}{c}{\makecell{Eyeglasses (Pre-trained)}}  & \multicolumn{2}{c}{\makecell{Bangs (Pre-trained)}}                   \\
     \cmidrule(lr){2-3} \cmidrule(lr){4-5}
                                     & F1 $\uparrow$ & Accuracy $\uparrow$ & F1 $\uparrow$ & Accuracy $\uparrow$  \\
     \midrule
     Non-adv classifier                 & 0.9789 & 97.93\% & 0.9218 & 92.50\% \\
     Reweighting                     & 0.9831 & 98.33\% & 0.9169 & 92.04\% \\ 
     Resample                        & 0.9811 & 98.15\% & 0.9100 & 91.51\% \\
     CutMix                          & 0.9813 & 98.16\% & 0.8993 & 90.63\% \\
     AutoAugment                     & 0.9840 & 98.42\% & 0.9052 & 91.11\% \\
     \midrule
     \ourmodel (ours)                & 0.9846 & 98.48\% & 0.8956 & 90.31\% \\
     \ourmodel + Reweight (ours)     & \textbf{0.9872} & \textbf{98.73\%} & \textbf{0.9285} & \textbf{93.04\%} \\
     \bottomrule
  \end{tabular}
 \caption{\label{tab:data_augmentation_appendix} Data augmentation for image classification in imbalanced datasets.}
\end{table*}

\begin{figure*}[h]
    \centering
    \includegraphics[width=0.7\textwidth]{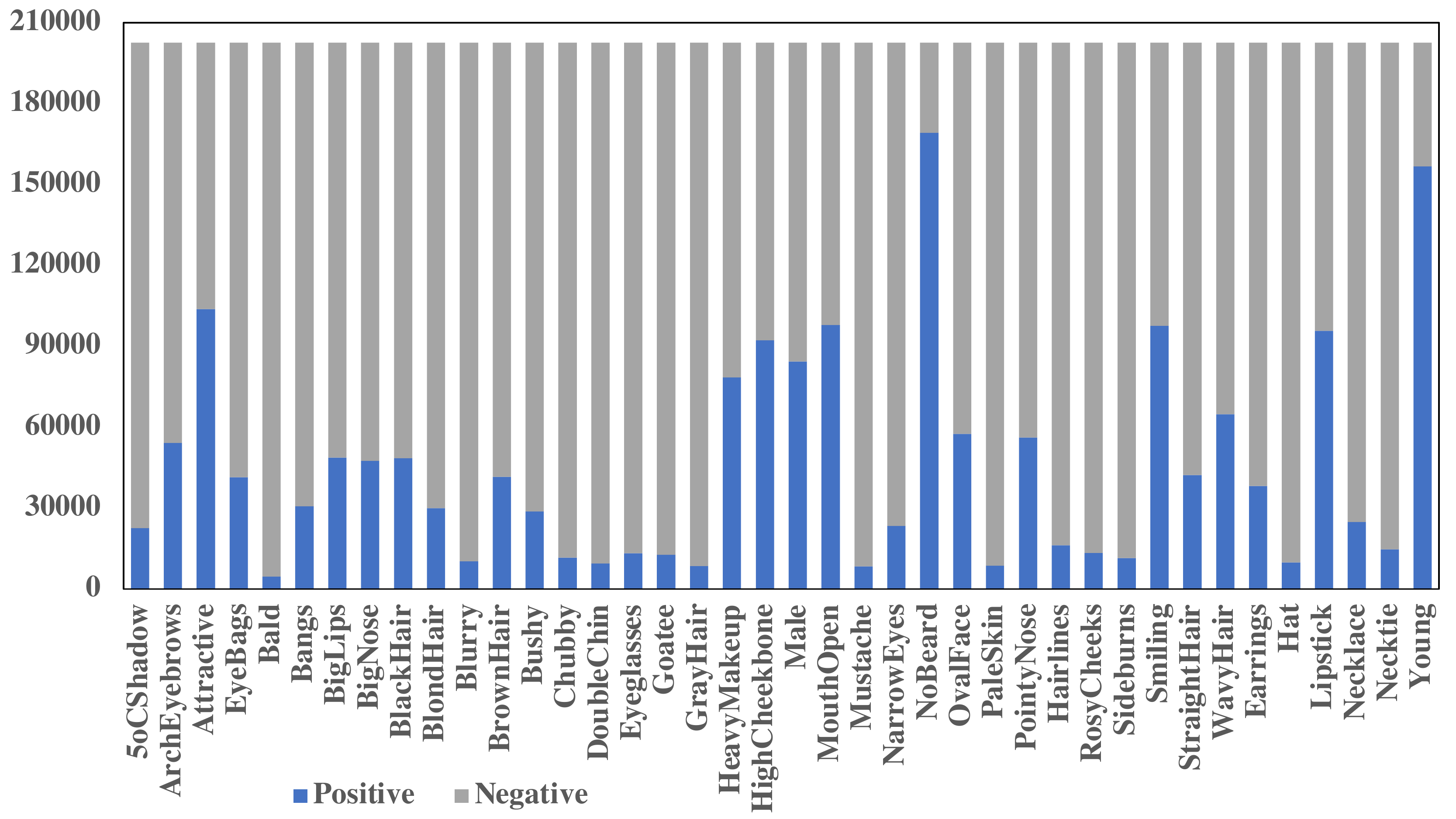}
    \caption{\label{fig:celeba_appendix_full} Label proportions of the whole CelebA dataset ($202599$ facial images).}
\end{figure*}

\newpage
\section{Attribute Sensitivity on Different Data Amount}
We use \ourmodel to evaluate the gender classifier and heavy-makeup classifiers trained with the setup in section 4.1. We take the first $17,000$ images from the CelebA test set and divide them into three sets of $10, 000$ images, $5, 000$ images, and $2, 000$ images by data index. Note that classifiers and $\mathcal{G}_\phi$ have never been trained on these images before. Table \ref{tab:SIA20_histogram_values}, \ref{tab:SIA20_histogram_values_cont}, \ref{tab:SIA20_histogram_values_2}, \ref{tab:SIA20_histogram_values_cont_2} show the results. We can see that the sensitivities are empirically consistent over different amounts of evaluation data. To take Table \ref{tab:SIA20_histogram_values}, \ref{tab:SIA20_histogram_values_cont} (gender classifier) as an example, we can see that with different amounts of data the shape of the histogram (Fig. \ref{fig:gender_data_amount_hist}) remains consistent. Note that the most sensitive attribute for attacking the gender classifier is consistently the lipstick. In the open-source version, we will release the code and instructions for running SIA on multiple setups of data.

\begin{table*}[h]
   \centering
   \small
   \begin{tabular}{@{}lcccccccc@{}} 
     \toprule
     & \makecell{Bald}     & \multicolumn{1}{c}{Bangs}&  \makecell{Black Hair}& \makecell{Blond Hair} &  \makecell{Brown Hair} & \multicolumn{1}{c}{Bushy} & \multicolumn{1}{c}{Eyeglasses} & \makecell{Mouth Open} \\
     \midrule
    $2000$ & 0.0434 & 0.0605 & 0.0405 & 0.0434 & 0.0545 & 0.0887 & 0.0803 & 0.0584 \\
    $5000$ & 0.0426 & 0.0580 & 0.0417 & 0.0445 & 0.0528 & 0.0879 & 0.0839 & 0.0551 \\
    $10000$ & 0.0427 & 0.0595 & 0.0409 & 0.0424 & 0.0532 & 0.0879 & 0.0838 & 0.0548 \\
     \bottomrule
   \end{tabular}
   \caption{\label{tab:SIA20_histogram_values} Detailed numerical values of each attribute sensitivity for the gender classifier using different amounts of data.}
   \vspace{-3mm}
\end{table*}

\begin{table*}[h]
   \centering
   \small
   \begin{tabular}{@{}lcccccc@{}} 
     \toprule
     & 
     \makecell{Mustache}& \makecell{No Beard} &  \makecell{Pale Skin} & \multicolumn{1}{c}{Young} & \multicolumn{1}{c}{Smiling} & \makecell{Lipstick}\\
     \midrule
    $2000$ & 0.0755 & 0.1495 & 0.0452 & 0.0647 & 0.0370 & \textbf{0.1585}  \\
    $5000$ & 0.0775 & 0.1493 & 0.0449 & 0.0656 & 0.0375 & \textbf{0.1586}\\
    $10000$ & 0.0775 & 0.1499 & 0.0458 & 0.0652 & 0.0359 & \textbf{0.1606} \\
     \bottomrule
   \end{tabular}
    \caption{\label{tab:SIA20_histogram_values_cont} (Cont.) Detailed numerical values of each attribute sensitivity for the gender classifier using different amounts of data.}
    \vspace{-3mm}
\end{table*}

\begin{table*}[h]
   \centering
   \small
   \begin{tabular}{@{}lccccccccc@{}} 
     \toprule
     & \makecell{Bald}     & \multicolumn{1}{c}{Bangs}&  \makecell{Black \\ Hair}& \makecell{Blond Hair} &  \makecell{Brown Hair} & \multicolumn{1}{c}{Bushy} & \multicolumn{1}{c}{Eyeglasses} & \multicolumn{1}{c}{Gender} & \makecell{Mouth Open} \\
     \midrule
    $2000$ & 0.0269 & 0.0772 & 0.0438 & 0.0540 & 0.0650 & 0.0742 & \textbf{0.1296} & 0.0446 & 0.0494 \\
    $5000$ & 0.0268 & 0.0774 & 0.0400 & 0.0528 & 0.0631 & 0.0792 & \textbf{0.1278} & 0.0446 & 0.0525 \\
    $10000$ & 0.0265 & 0.0783 & 0.0419 & 0.0563 & 0.0645 & 0.0740 & \textbf{0.1294} & 0.0419 & 0.0494 \\
     \bottomrule
   \end{tabular}
   \caption{\label{tab:SIA20_histogram_values_2} (Cont.) Detailed numerical values of each attribute sensitivity for the heavy-makeup classifier using different amounts of data.}
   \vspace{-3mm}
\end{table*}

\begin{table*}[h]
   \centering
   \small
   \begin{tabular}{@{}lcccccc@{}} 
     \toprule
     & 
     \makecell{Mustache}& \makecell{No Beard} &  \makecell{Pale Skin} & \multicolumn{1}{c}{Young} & \multicolumn{1}{c}{Smiling} & \makecell{Lipstick}\\
     \midrule
    $2000$ & 0.0740 & 0.1244 & 0.0291 & 0.0691 & 0.0371 & 0.1016 \\
    $5000$ & 0.0772 & 0.1213 & 0.0298 & 0.0651 & 0.0408 & 0.1015\\
    $10000$ & 0.0790 & 0.1218 & 0.0288 & 0.0679 & 0.0368 & 0.1037\\
     \bottomrule
   \end{tabular}
   \caption{\label{tab:SIA20_histogram_values_cont_2} (Cont.) Detailed numerical values of each attribute sensitivity for the heavy-makeup classifier using different amounts of data.}
   \vspace{-3mm}
\end{table*}

\begin{figure*}[h]
    \centering
    \includegraphics[width=\linewidth]{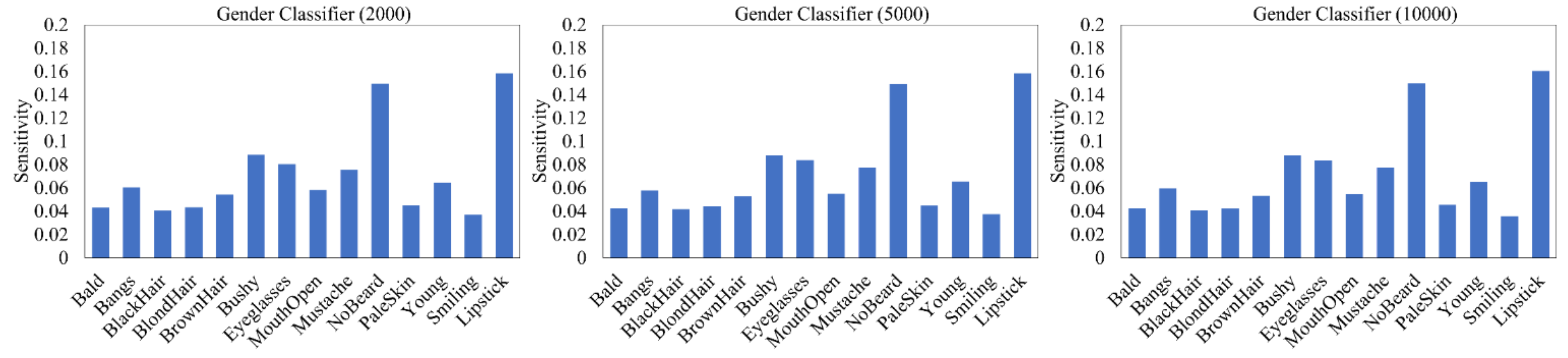}
    \caption{\label{fig:gender_data_amount_hist} Histogram visualization of attribute sensitivities (under different \ourmodel data amount) for the gender classifier.}
\end{figure*}
\newpage
\section{Attack Effectiveness}
We implement SPT \cite{2019semanticadversarial}, CW \cite{2017CWAttack}, and Face-Manifold (FM) \cite{Li_2021_CVPR} attacks and evaluate the attack success rates (ASR) on our facial eyeglass classifier. The classifier is trained with the setup mentioned in section 4.1. For all listed setups (unless otherwise stated), the images for evaluation is the first $2,000$ images from CelebA test set.

In SPT attack, we use the attribute space consisting the same $15$ attributes as SIA. The optimizer is RMSProp with two learning rates $\eta = {0.25}/{255}$ and $\eta = {4}/{255}$. Table \ref{tab:SPT_ASR} shows the ASR with different attack iterations. Under the same setup of $200$ iterations with $\eta = {0.25}/{255}$, our SIA \textit{attribute-only} ASR (in section 4.3 ablation study table) is 32.67\% which outperforms SPT. This shows that the use of signed gradient to update the attribute space, which stabilizes the optimization, can improve the attack effectiveness.

\begin{table}[!h]
   \centering
   \begin{tabular}{@{}lccccccccccc@{}} 
     \toprule
     Iterations & \makecell{2} & \makecell{5} & \makecell{10} & \makecell{15} & \makecell{20} & \makecell{50} & \makecell{100} & \makecell{150} & \makecell{200}   \\
     \midrule
    $\eta = \frac{0.25}{255}$ & 0.3\% & 0.3\% & 0.4\% & 0.7\% & 0.9\% & 3.1\% & 12.3\% & 19.4\% & \textbf{22.9\%}  \\
    $\eta = \frac{4}{255}$ & 1.2\% & 5.0\% & 14.6\% & 21.7\% & 25.4\% & 28.3\% & 29.6\% & \textbf{30.3\%} & 29.7\%  \\
     \bottomrule
   \end{tabular}
   \caption{\label{tab:SPT_ASR} ASR for SPT attack with different step size $\eta$ on eyeglasses classifier}
\end{table}

In CW attack, we fix the attack iteration same as \ourmodel's $200$ and evaluate the ASR under different box-constrain parameters. Table \ref{tab:CW_ASR} shows that by relaxing the box-constrain, the ASR can hit to 52.6\% which is higher than our PGD baseline of 49.85\%. The default setting of SIA where image and attribute spaces are co-updated has an ASR of 68.20\% which are much more effective than CW and PGD. Note that during SIA's adversarial optimization, we can obtain attribute sensitivity which provides intuitive model interpretation to users. Pure image space attacks cannot support such features.

\begin{table}[!h]
   \centering
   \begin{tabular}{@{}lccccccccc@{}} 
     \toprule
     \makecell{Box Contrain} & \makecell{0.10} & \makecell{0.25} & \makecell{0.50} & \makecell{0.75} & \makecell{1.0} & \makecell{1.5} & \makecell{2.0} & \makecell{5.0} & \makecell{10.0}  \\
     \midrule
     ASR & 9.1\% & 16.8\% & 25.0\% & 29.4\% & 31.9\% & 35.1\% & 37.2\% & 45.4\% & \textbf{52.6\%} \\
     \bottomrule
   \end{tabular}
   \caption{\label{tab:CW_ASR} ASR for CW attack with different box-constrains setup on eyeglasses classifier}
\end{table}

In FM attack, we follow the setup as specified in the original paper to make sure that we can re-implement the high ASR reported in their paper. We sample $2,000$ images from the style space to experiment on different settings of $\epsilon_{1}$ (style step size) and  $\epsilon_{2}$ (noise step size). Table \ref{tab:FM_ASR} shows the ASR of different $\epsilon_{1}$ and  $\epsilon_{2}$ settings. We find out that noise vectors have a superior effect on flipping the prediction of our eyeglasses ResNet classifier. With increasing the strength of injected noises during the generation, the image quality will significantly decrease. Note that SIA and PGD can also achieve similar ASR (99.99\% and 99.94\% correspondingly) by relaxing the image space constraint.

\begin{table}[!h]
  \centering
  \begin{tabular}{@{}lcccccc@{}}
     \toprule
                                     & \makecell{$\epsilon_{2}$ = 0 (no noise)}   
                                     & \makecell{$\epsilon_{2}$ = 0.01}
                                     & \makecell{$\epsilon_{2}$ = 0.02}
                                     & \makecell{$\epsilon_{2}$ = 0.03}
                                     & \makecell{$\epsilon_{2}$ = 0.04}
                                     & \makecell{$\epsilon_{2}$ = 0.05}
                                     \\
     \midrule
     $\epsilon_{1}$ = 0.004                  & 2.5\% & 70.8\% & 96.6\% &  99.7\% & 100.0\% &  99.9\% \\
     $\epsilon_{1}$ = 0.01                   & 2.2\% & 19.3\% & 55.9\% & 78.8\% & 91.9\% &  97.3\% \\
     $\epsilon_{1}$ = 0.05             & 0.3\% & 1.9\% & 3.95\% & 8.2\% & 19.3\% &  20.5\% \\
     $\epsilon_{1}$ = 0.1                    & 0.2\% & 0.8\% & 1.3\% & 2.9\% & 7.0\% &  12.1\% \\
     \bottomrule
  \end{tabular}
  \caption{\label{tab:FM_ASR} ASR for FM attack on eyeglasses classifier}
\end{table}

In the open source version, we will release the code and instructions on how to set up hyperparameters, schedule baseline experiments, and run code for comparisons.
}

\end{document}